\documentclass{article}

\let\originaladdcontentsline\addcontentsline

\usepackage[preprint]{colm2026_conference}

\usepackage{hyperref}
\usepackage{url}
\usepackage{booktabs}
\usepackage{amsfonts}
\usepackage{nicefrac}
\usepackage{microtype}
\usepackage[table]{xcolor}
\definecolor{SkyBlue}{rgb}{0.53,0.81,0.92}
\providecommand{\MethodName}{EvoStreaming}
\providecommand{\VarSty}[1]{\textnormal{\ttfamily\color{blue!90!black}#1}\unskip}
\usepackage{graphicx}
\usepackage{wrapfig}
\usepackage{mathtools}
\usepackage{subcaption}
\usepackage{multirow}
\usepackage{enumitem}
\usepackage{amsmath}
\usepackage{amssymb}
\usepackage{pifont}

\newcommand{\xmark}{{\ding{55}}}
\usepackage{amsthm}
\usepackage{algorithm}
\usepackage{algpseudocode}
\usepackage{array}
\usepackage{tcolorbox}
\tcbuselibrary{breakable,skins}
\usepackage{varwidth}
\usepackage{fontawesome}
\usepackage{caption}
\usepackage{titletoc}
\usepackage[capitalize,noabbrev]{cleveref}
\usepackage{makecell}

\usepackage{lineno}

\definecolor{darkblue}{rgb}{0, 0, 0.5}
\hypersetup{colorlinks=true, citecolor=darkblue, linkcolor=darkblue, urlcolor=darkblue}

\definecolor{darkpink}{RGB}{255, 20, 147}

\makeatletter
\def\input@path{{../}{./}}
\makeatother
\graphicspath{{../}{./}}

\title{EvoStreaming: Your Offline Video Model Is a Natively Streaming Assistant}

\author{%
Zichen Wen$^{1}$\thanks{Equal Contribution.} \quad
Boxue Yang$^{1}$\footnotemark[1] \quad
Junlong Ke$^{2}$\footnotemark[1] \quad
Jiajie Huang$^{1}$ \\
\textbf{Chenfei Liao$^{3}$} \quad
\textbf{Junxi Wang$^{1,4}$} \quad
\textbf{Xuyang Liu$^1$} \quad
\textbf{Linfeng Zhang$^{1}$\thanks{Corresponding author: zhanglinfeng@sjtu.edu.cn.}} \\
\textsuperscript{1}EPIC Lab, Shanghai Jiao Tong University
\quad
\textsuperscript{2}Tsinghua University \\
\textsuperscript{3}The Hong Kong University of Science and Technology (Guangzhou)
\textsuperscript{4}Fudan University
\vspace{4pt} \\
\faGithub \, \textbf{Code:} \href{https://github.com/BoxueYang/EvoStreaming}{\textcolor{darkpink}{https://github.com/BoxueYang/EvoStreaming}}
}

\begin{document}

\ifcolmsubmission
\linenumbers
\fi

\maketitle
\lhead{Preprint.}

\begin{abstract}

Streaming video understanding demands more than watching longer videos: assistants must decide \emph{when} to speak in real time, balancing responsiveness against verbosity. Yet most video-language models (VideoLLMs) are trained for offline inference, and existing streaming benchmarks externalize this timing decision to the evaluator. We address this gap with \textbf{RealStreamEval}, a frame-level multi-turn evaluation protocol that exposes models to sequential observations and penalizes unnecessary responses. Under this protocol, we observed that strong offline VideoLLMs retain useful visual understanding but lack an interaction policy for deciding when to respond. Motivated by this observation, we propose \textbf{EvoStreaming}, a self-evolved streaming adaptation framework in which the base model itself acts as data generator, relevance annotator, and roll-out policy to synthesize streaming trajectories without external supervision. With only $1{,}000$ self-generated samples ($139\times$ less than the leading streaming instruction-tuning approach) and no architectural changes, EvoStreaming consistently improves the overall RealStreamEval score by up to $10.8$ points across five open VideoLLM backbones (Qwen2/2.5/3-VL, InternVL-3.5, MiniCPM-V4.5) while largely preserving offline video performance. These results suggest that data-efficient interaction tuning is a practical path for adapting existing VideoLLMs to streaming assistants.
\end{abstract}

\section{Introduction}
\label{sec:intro}

\begin{wrapfigure}{r}{0.52\textwidth}
    \vspace{-1.2em}
    \centering
    \includegraphics[width=\linewidth]{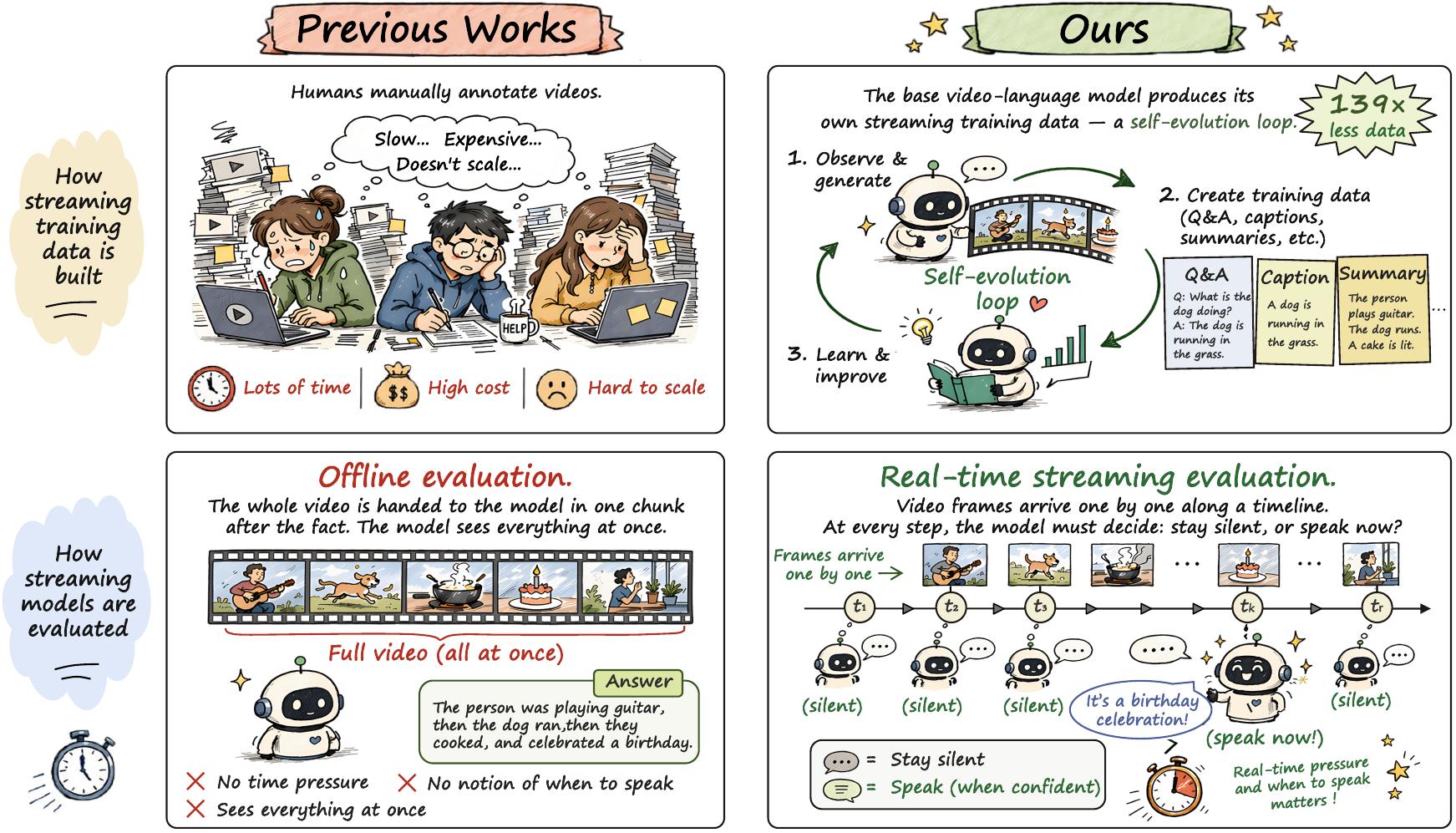}
    \caption{\textbf{EvoStreaming} lets a VideoLLM teach itself \emph{when} to speak in a video stream (top), evaluated by \textbf{RealStreamEval}, which scores both correctness and response timing (bottom).}
    \label{fig:teaser}
    \vspace{-1.2em}
\end{wrapfigure}

Real-time video is becoming a primary interface for interacting with the world.
From live streaming and telepresence to embodied agents~\citep{yang2026efficientvla} and wearable assistants~\citep{wen2025ai}, users increasingly expect systems to watch continuously, answer questions on demand, and proactively surface important events under strict latency~\citep{liu2025shifting,wang2025proactivevideoqacomprehensivebenchmarkevaluating,yan2026proactvl} and energy budgets.
However, as summarized in Figure~\ref{fig:teaser}, most video-language models (VideoLLMs) are still developed in an \emph{offline} setting that assumes full-video access before any response is produced~\citep{wang2024qwen2,Qwen3-VL,chen2024internvl,zhang2024llavavideo,team2026kimi}.

Streaming is not simply ``longer video''.
Video frames arrive sequentially, and the assistant must decide not only \emph{what} to say but also \emph{when} to speak.
This ``when-to-respond'' decision is central to autonomy and efficiency: responding too late misses events, while responding too often becomes verbose, distracts users, and wastes compute, especially on resource-constrained edge devices such as smart glasses.
Recent work has therefore begun to study \emph{streaming video understanding}~\citep{xia2025streaming,wang2025accelerating}, with approaches ranging from memory mechanisms for long-stream retention~\citep{zhang2024flashvstream}, to explicit timing control modules~\citep{qian2025dispider}, continuous-input online architectures~\citep{chen2024videollm}, and token compression~\citep{liu2025video,liu2026global,xiong2025prune2drive,yao2025timechatonline,wen2025token,wen2025stop,wen2026efficient,chen2025variation,ke2026flashunified,li2026catp} for latency reduction.

Despite this progress, streaming video understanding faces two practical bottlenecks: evaluation misalignment and streaming adaptation cost.

\textbf{\emph{Existing streaming evaluation protocols are often misaligned with real deployment.}}
Popular streaming benchmarks either feed video as growing chunks for offline inference, or schedule external timing prompts at fixed intervals~\citep{li2025ovobench,lin2024streamingbench}.
Both designs shift the timing decision to the evaluator and score only answer accuracy, leaving response frequency and verbosity unmeasured.

\textbf{\emph{Streaming adaptation is costly in practice.}}
Architecture-oriented methods add streaming-specific modules and re-train for online interaction~\citep{zhang2024flashvstream,qian2025dispider,chen2024videollm}, which incurs substantial engineering effort and is hard to retrofit onto deployed backbones.
Data-scaling methods adapt offline models with large-scale curated or synthetic streaming instructions~\citep{yao2025timechatonline,xia2025streaming}, which are expensive to produce, verify, and maintain.
Both routes raise the cost of bringing a strong offline backbone online.

We address the first bottleneck with \textbf{RealStreamEval}, a frame-level multi-turn evaluation protocol that feeds frames sequentially and lets the model itself decide, at each step, whether to remain silent or produce a response, thereby better mirroring realistic deployment where a streaming assistant observes a live feed and must autonomously choose when to intervene.
RealStreamEval scores not only answer correctness at each response step, but also whether the response timing is appropriate under sequential observations, and it explicitly penalizes mistimed or excessive responses through a nonlinear verbosity term.
Under this stricter protocol, we find that strong offline VideoLLMs largely retain visual understanding, but lack a reliable interaction policy for deciding when to speak.

Informed by this observation, we address the remaining streaming adaptation obstacle with \textbf{EvoStreaming}, a self-evolved streaming adaptation framework.
Unlike prior adaptation methods that redesign model architectures or scale to large amounts of streaming instruction data, EvoStreaming asks the base video-language model itself to act in three roles: \emph{data generator} (producing streaming questions), \emph{relevance annotator} (labeling segment-level visual evidence), and \emph{roll-out policy} (turning the relevance labels into a causal \textsc{Silent}/respond trajectory).
The trajectories then fine-tune the model itself, teaching it two complementary skills: when to surface a response as visual evidence emerges, and when to remain silent before any evidence is available.
With only $1{,}000$ self-generated samples ($139\times$ less than the leading streaming instruction-tuning approach) and no architectural modification, EvoStreaming improves the overall RealStreamEval score by up to $10.8$ points across five open VideoLLM backbones (Qwen2/2.5/3-VL, InternVL-3.5, MiniCPM-V4.5), while largely preserving offline video performance.

Taken together, this work makes three main contributions:
\begin{itemize}[leftmargin=10pt, topsep=0pt, itemsep=1pt, partopsep=1pt, parsep=1pt]
    \item We introduce \textbf{RealStreamEval}, a deployment-faithful protocol that jointly evaluates answer correctness, response timing, and response economy via an explicit verbosity penalty, enabling a stricter and more realistic assessment than prior chunked or polling-based setups.
    \item We propose \textbf{EvoStreaming}, a self-evolved, architecture-preserving streaming adaptation framework, where the base model itself acts as data generator, relevance annotator, and roll-out policy to synthesize causal \textsc{Silent}/respond trajectories, substantially reducing adaptation overhead without adding model modules.
    \item EvoStreaming improves RealStreamEval by up to $10.8$ using only $1{,}000$ self-generated samples on five open VideoLLM backbones ($139\times$ less data than prior streaming video understanding methods), while largely preserving offline video understanding performance.
\end{itemize}

\section{Related Work}
\label{sec:related}

\noindent
\textbf{Video Large Language Models.}
Recent multimodal LLMs~\citep{kimivl,team2026longcat,wen2026innovator} have rapidly extended visual instruction tuning~\citep{liu2023llava,dai2023instructblip} from static images to video, with open backbones such as Qwen-VL~\citep{wang2024qwen2,Qwen3-VL}, Intern-VL~\citep{chen2024internvl}, Kimi K2.5~\citep{team2026kimi}, and LLaVA-Video~\citep{zhang2024llavavideo} achieving strong long-form video understanding~\citep{dai2026webvr}. These models are typically trained under offline assumptions: the full video, or a preselected clip, is available before any response is produced. As a result, they do not natively handle streaming interaction, where frames arrive sequentially and the model must decide, at each step, whether to remain silent or respond.

\noindent
\textbf{Streaming Video Understanding.}
Recent work on real-time video interaction follows two routes. \emph{Architecture-oriented} methods, such as Flash-VStream~\citep{zhang2024flashvstream}, Dispider~\citep{qian2025dispider}, VideoLLM-online~\citep{chen2024videollm}, StreamBridge~\citep{wang2025streambridge}, and PEARL~\citep{zheng2026pearl}, introduce memory mechanisms, perception-decision-reaction modules, dedicated control tokens, or activation models to regulate response timing; all require model-specific engineering and are hard to retrofit onto backbones. \emph{Data-driven} methods, such as TimeChat-Online~\citep{yao2025timechatonline}, adapt offline models via large-scale streaming instruction tuning, which is expensive to produce, verify, and maintain. EvoStreaming differs from both: it introduces no streaming-specific module and uses $139\times$ less supervision than the leading data-driven approach, by letting the base model itself synthesize the streaming trajectories on which it is fine-tuned.

\noindent
\textbf{Evaluation for Streaming Video Understanding.}
Streaming evaluation must capture both visual correctness and response timing. OVO-Bench~\citep{li2025ovobench} provides a rich task taxonomy but processes history as chunked offline inputs. StreamingBench~\citep{lin2024streamingbench} adopts a polling strategy that asks the evaluator, rather than the model, when to respond. Both designs are reproducible but partially externalize the timing decision. RealStreamEval reorganizes the same style of tasks into a frame-level multi-turn protocol where the model itself decides when to speak, and adds an explicit verbosity penalty that captures response-frequency costs left unmeasured by prior protocols~\citep{wen2025ai}. Table~\ref{tab:method_comparison} summarizes the key design differences among streaming adaptation methods.

\begin{table*}[!t]
  \centering
  \setlength{\tabcolsep}{4pt}
  \renewcommand{\arraystretch}{1.15}
  \caption{\textbf{Qualitative comparison of streaming video adaptation methods.} \MethodName{} requires no architectural modification or external inference module, uses $139\times$ less supervision than the leading data-driven baseline, needs no human annotation, and generalizes across five diverse backbones without per-model engineering.}
  \label{tab:method_comparison}
  \resizebox{\linewidth}{!}{%
  \begin{tabular}{@{}lccccc@{}}
  \toprule
  \textbf{Method} & \textbf{Arch.\ Modification} & \textbf{Inference Module} & \textbf{Streaming Data} & \textbf{Annotation} & \textbf{\#Backbones} \\
  \midrule
  Flash-VStream~\citep{zhang2024flashvstream}       & STAR memory            & \xmark            & --     & Human    & 1 \\
  VideoLLM-online~\citep{chen2024videollm}           & Attn.\ mask + EOS head & Async pipeline    & >134K  & Human    & 1 \\
  Dispider~\citep{qian2025dispider}                  & PDR modules            & \xmark            & --     & Human    & 1 \\
  StreamBridge~\citep{wang2025streambridge}           & Mem.\ buffer + compress.  & 0.5B activ.\ model & 120K   & Human    & 3 \\
  TimeChat-Online~\citep{yao2025timechatonline}       & \xmark                 & \xmark            & 139K   & Human    & 1 \\
  \midrule
  \rowcolor{SkyBlue!20}
  \MethodName{} (ours)                                & \xmark                 & \xmark            & \textbf{1K (self-gen)} & \textbf{Self-gen} & \textbf{5} \\
  \bottomrule
  \end{tabular}%
  }
\vspace{-1.5em}
\end{table*}

\section{Methodology}
\label{sec:method}

\subsection{Motivation}
\label{sec:motivation}
RealStreamEval reveals that strong offline VideoLLMs retain useful visual understanding but lack a reliable policy for deciding when to speak. Rather than relearning visual semantics, streaming adaptation can focus on the lower-dimensional problem of interaction timing. EvoStreaming materializes this insight by preserving the base architecture and shifting the adaptation burden to a self-evolved timing policy: the base model itself generates $1,000$ streaming trajectories that supervise \textsc{Silent}/respond decisions, with no external annotator or architectural modification.

\begin{figure*}[!t]
    \centering
    \includegraphics[width=\linewidth]{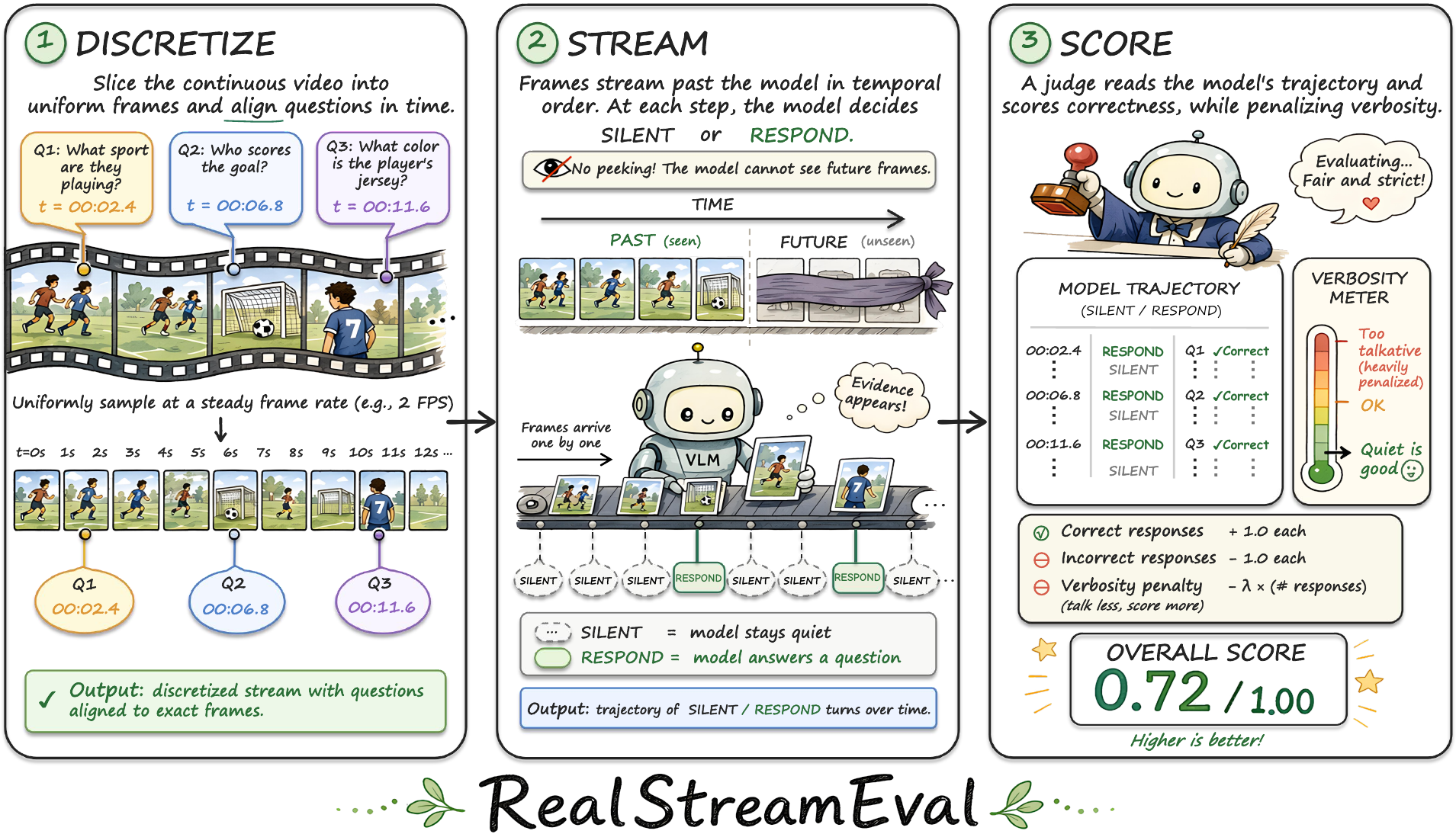}
    \caption{\textbf{Overview of RealStreamEval.} We first align questions to sampled frames, then run strict online inference where the model decides when to respond, and finally score responses with a verbosity penalty to discourage redundant outputs.}
    \label{fig:RealStreamEval}
    \vspace{-1.2em}
\end{figure*}

\subsection{RealStreamEval: Evaluating When to Speak, Not Just What to Say}
\label{sec:realestreameval}
Many existing streaming video evaluation frameworks still simplify parts of the streaming decision process.
Take OVO-Bench as an example: it evaluates by trimming the video from the beginning up to the question timestamp, then feeding the entire clip to the model in one shot; StreamingBench, in active-response tasks, adopts a polling strategy, querying the model every second: ``Is it time to respond?''. These choices make evaluation reproducible, but they partially shift the timing decision to the evaluator. As illustrated in Figure~\ref{fig:RealStreamEval}, RealStreamEval instead evaluates the model under sequential observations and asks it to decide when to respond.

\paragraph{Streaming as a sequential decision.}
\label{sec:frame_dialogue}
RealStreamEval reformulates streaming video understanding as a sequential decision process. Frames arrive at a fixed rate (we use $0.5$ fps), and a user query $q_t$ may be issued at any timestep $t \in \{1, \ldots, T\}$ (with $q_t = \emptyset$ if no new query is issued). At each step the model selects an action
\begin{equation}
\label{eq:action_space}
a_t \;\in\; \mathcal{A} \;=\; \{\textsc{Silent}\} \,\cup\, \mathcal{R},
\end{equation}
where $\mathcal{R}$ denotes the space of natural-language responses. The decision must be \emph{causal}: it depends only on the history available up to $t$,
\begin{equation}
\label{eq:history}
\mathcal{H}_t \;=\; \bigl(v_{1:t},\; q_{1:t},\; a_{1:t-1}\bigr), \qquad a_t \;=\; \pi\bigl(\mathcal{H}_t\bigr),
\end{equation}
so future frames $v_{t+1}, \ldots, v_T$ are inaccessible at time $t$. This is in contrast to standard offline inference, $a^{\star} = f_{\text{offline}}(v_{1:T}, q)$, which conditions on the entire stream and therefore sidesteps the core streaming challenge of acting under partial information.


The frame-level dialogue alone does not capture a key deployment constraint: a streaming assistant should not speak unnecessarily, since response frequency directly affects user experience and device energy. RealStreamEval addresses this by combining timed answer matching with a multiplicative verbosity penalty.

\paragraph{What we track.}
Let $\mathcal{T}^{\star} \subseteq \{1, \ldots, T\}$ denote the ground-truth response timestamps of a sample, with reference answers $a^{\star}_{t^{\star}}$ at each $t^{\star} \in \mathcal{T}^{\star}$. Let $\mathcal{T} = \{t : a_t \neq \textsc{Silent}\}$ denote the timestamps at which the model produces a response. We define the \emph{turn answer rate} as $R_{\text{ans}} = |\mathcal{T}|/T$.

\paragraph{Beyond accuracy: rewarding silence.}
\label{sec:verbosity_penalty}
We adopt a strict \textbf{multiplicative decay} strategy that prevents a verbose model from inflating its score by responding at every turn. The overall sample score combines response quality, timing accuracy, and the verbosity multiplier:
\begingroup
\setlength{\abovedisplayskip}{10pt}
\setlength{\belowdisplayskip}{10pt}
\setlength{\abovedisplayshortskip}{10pt}
\setlength{\belowdisplayshortskip}{10pt}
\begin{equation}
\label{eq:score}
S \;=\; \underbrace{\frac{1}{|\mathcal{T}^{\star}|} \sum_{t^{\star} \in \mathcal{T}^{\star}} \max_{\substack{t \in \mathcal{T} \\ |t - t^{\star}| \le \delta}} \mathcal{J}\bigl(a_t,\, a^{\star}_{t^{\star}}\bigr)}_{\text{response quality}} \;\cdot\; \underbrace{\mathcal{M}(R_{\text{ans}})}_{\text{verbosity multiplier}} \;-\; \underbrace{P_{\text{early}}}_{\text{premature penalty}},
\end{equation}
\endgroup
Here $\delta$ defines temporal tolerance for matching predicted and ground-truth response timestamps, $\mathcal{J}(\cdot, \cdot) \in [0,1]$ denotes a judge function that assesses answer quality, and we adopt the convention $\max_{\emptyset}\mathcal{J} = 0$ when no model response falls within the matching window of $t^{\star}$. Recognizing that models may adopt varying response formats and strategies, we employ an LLM-as-judge approach: Qwen3-VL-235B-A22B-Instruct~\citep{Qwen3-VL} served via vLLM~\citep{kwon2023efficient} evaluates whether the predicted response $a_t$ correctly addresses the query, accounting for semantic equivalence rather than exact string matching. This flexible assessment ensures fair comparison across diverse response styles.

\paragraph{Two regimes, two penalty profiles.}
The penalty terms instantiate task-specific constraints. \emph{Forward Active Responding} (FAR) tasks require silence before visual evidence emerges, so we set $P_{\text{early}} = 0.1$ and a piecewise multiplier
\begin{equation}
\label{eq:m_step}
\mathcal{M}_{\text{FAR}}(R_{\text{ans}}) \;=\;
\begin{cases}
1.0 & R_{\text{ans}} < 0.4, \\
\max\!\big(1 - 0.2\,\lfloor 5R_{\text{ans}}\rfloor,\, 0.2\big) & R_{\text{ans}} \ge 0.4,
\end{cases}
\end{equation}
that strictly penalizes high-frequency responding. For \emph{Real-Time Visual Perception} and \emph{Backward Tracing} tasks, ``early'' is undefined, so we set $P_{\text{early}} = 0$, collapse the matching window to the current frame ($\delta = 0$), and use $\mathcal{M}$ as a repetition penalty: $\mathcal{M}_{\text{rep}} = 0.5$ if the same answer is emitted across consecutive turns and $1.0$ otherwise.

We provide additional details on RealStreamEval's judge design and protocol scope in Appendix~\ref{app_sec:reliability_scope}.

\begin{figure*}[!t]
    \centering
    \includegraphics[width=\linewidth]{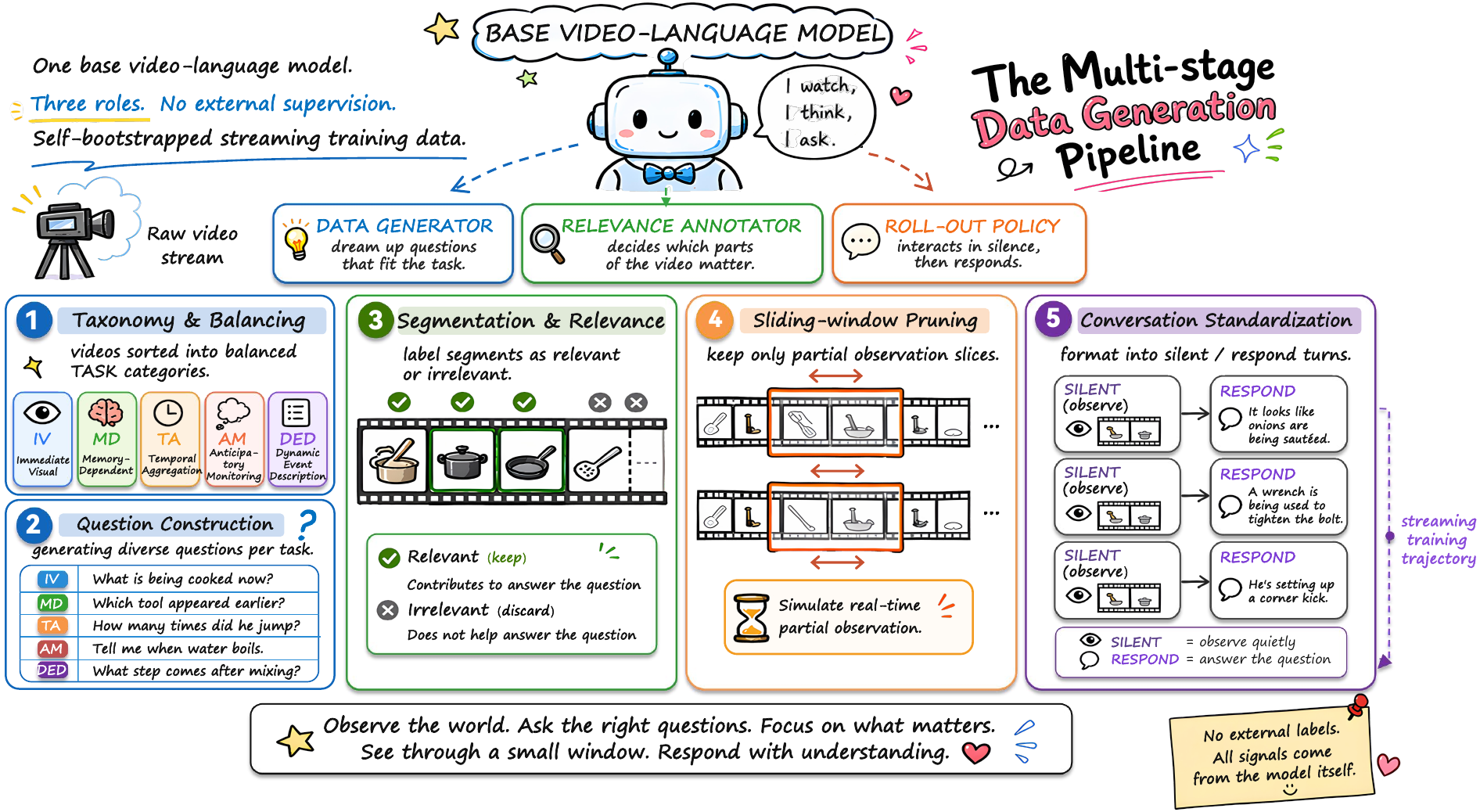}
    \vspace{-1.5em}
\caption{\textbf{Multi-stage data generation pipeline.}
Raw videos are converted into streaming training data through five stages:
taxonomy balancing, type-consistent question construction, segmentation with relevance annotation, sliding-window pruning under partial observations, and conversation standardization.
}
\label{fig:data_pipeline}
\end{figure*}

\subsection{EvoStreaming: Letting the Model Teach Itself When to Wait}
\label{sec:evostreaming}

%
%

RealStreamEval suggests that strong offline VideoLLMs already provide a useful visual foundation; what they lack is an \emph{interaction policy} for deciding when to speak. EvoStreaming therefore preserves the base architecture and shifts the burden of streaming adaptation from visual recognition (already learned during offline pre-training) to interaction timing. We materialize this idea by letting the base model itself play generator, annotator, and roll-out policy, producing streaming trajectories that supervise \emph{when} to respond.

\paragraph{The model as its own annotator.}
Given a raw video $\mathcal{V}$ uniformly segmented into $S$ intervals $\{s_1,\ldots,s_S\}$, the base model $\mathcal{M}$ first assigns a task type $c$ from five categories that span reactive responses and patient observation: \textit{Immediate Visual} (IV), \textit{Memory-Dependent} (MD), \textit{Temporal Aggregation} (TA), \textit{Anticipatory Monitoring} (AM), and \textit{Dynamic Event Description} (DED); their definitions and loose correspondence to the OVO-Bench evaluation modes are given in Appendix~\ref{sec:task_taxonomy}. Conditioned on $c$, $\mathcal{M}$ self-generates $K$ type-consistent questions $\mathcal{Q}=\{q_1,\ldots,q_K\}$ and a binary \emph{relevance matrix} $\mathbf{A}\in\{\textsc{I},\textsc{R}\}^{S\times K}$ that marks when visual evidence becomes available for each question. A causal sliding-window roll-out then converts $\mathbf{A}$ into a trajectory of $\textsc{Silent}$/response actions, which are aggregated into a conversational dataset $\mathcal{D}$ used to fine-tune $\mathcal{M}$. The full procedure, summarized in Algorithm~\ref{alg:evo} and illustrated in Fig.~\ref{fig:data_pipeline}, can be repeated for $I$ iterations so that later rounds bootstrap from a model with sharper timing behavior.

\begin{algorithm}[!t]
\caption{\textbf{EvoStreaming.} Each stage reuses the base model $\mathcal{M}$ in a different role, so no external annotator is required.}
\label{alg:evo}
\begin{algorithmic}[1]
\Require base model $\mathcal{M}^{(0)}$, video pool $\mathbb{V}$, iterations $I$
\Ensure adapted model $\mathcal{M}^{(I)}$
\For{$i = 0, \ldots, I-1$}
  \State $\mathcal{D} \gets \emptyset$
  \For{$\mathcal{V} \in \mathbb{V}$}
    \State $c \gets \mathcal{M}^{(i)}_{\mathrm{cls}}(\mathcal{V})$ \textcolor{blue}{\Comment{Stage 1: task-aware taxonomy}}
    \State $\mathcal{Q} \gets \mathcal{M}^{(i)}_{\mathrm{gen}}(\mathcal{V}, c)$ \textcolor{blue}{\Comment{Stage 2: self-questioning}}
    \State $\mathbf{A} \gets \bigl[\mathcal{M}^{(i)}_{\mathrm{rel}}(s_t, q_k)\bigr]_{t,k}$ \textcolor{blue}{\Comment{Stage 3: temporal grounding}}
    \For{$t = 1, \ldots, S$} \textcolor{blue}{\Comment{Stage 4: causal roll-out}}
      \If{$\mathbf{A}[t,k]=\textsc{I}$ for all $k$}
        \State $a_t \gets \textsc{Silent}$
      \Else
        \State pick $k^{\star}$ with $\mathbf{A}[t,k^{\star}]=\textsc{R}$;\;\; $a_t \gets \mathcal{M}^{(i)}(\mathcal{V}_{1:t}, q_{k^{\star}}) \in \mathcal{R}$
      \EndIf
      \State $\mathcal{D} \gets \mathcal{D} \cup \{(\mathcal{V}_{1:t},\, q_{1:t},\, a_t)\}$
    \EndFor
  \EndFor
  \State $\mathcal{M}^{(i+1)} \gets \textsc{Finetune}(\mathcal{M}^{(i)}, \mathcal{D})$ \textcolor{blue}{\Comment{Stage 5: self-evolution step}}
\EndFor
\end{algorithmic}
\end{algorithm}

Every stage reuses $\mathcal{M}$ in a different role, so the supervision is aligned with the model's own representation space and requires no external annotator. In our experiments, $I\in\{1,2,3\}$ outer iterations and $|\mathcal{D}|=1{,}000$ trajectories suffice to teach the timing policy across five backbones, using \textbf{139$\times$} less data than TimeChat-Online. Here $I=1$ corresponds to a single round of fine-tuning on self-generated data without iterative refinement, while $I\ge 2$ engages the self-evolution loop. The category distribution of $\mathcal{D}$ is visualized in Appendix Fig.~\ref{fig:data_distribution}; we discuss the quality and failure modes of self-generated supervision in Appendix~\ref{sec:self_data_quality}.


\section{Experiments}
\label{sec:experiments}

\begin{table*}[!t]
  \centering
  \setlength{\tabcolsep}{3pt}
  \renewcommand{\arraystretch}{1.35}
  \caption{\textbf{Results on OVO-Bench under RealStreamEval.} We report accuracy (\%) over three dimensions and 12 sub-tasks. All models are evaluated at 0.5 fps under RealStreamEval, which reuses OVO-Bench tasks and QA data with stricter streaming constraints and verbosity penalization. For all results here, each frame is processed with 768 visual tokens.}
  \label{tab:main_exp}
  \resizebox{\textwidth}{!}{%
  \begin{tabular}{@{}lccccccccccccccccc@{}}
  \toprule
  \multicolumn{1}{l|}{\multirow{2}{*}{Model}} &
  \multicolumn{1}{c|}{\multirow{2}{*}{\# Frames}} &
  \multicolumn{7}{c|}{Real-Time Visual Perception} &
  \multicolumn{4}{c|}{Backward Tracing} &
  \multicolumn{4}{c|}{Forward Active Responding} &
  \multicolumn{1}{c}{\multirow{2}{*}{Overall Avg.}} \\
  \cmidrule(l){3-17}
  \multicolumn{1}{l|}{} & \multicolumn{1}{c|}{} &
  OCR & ACR & ATR & STU & FPD & \multicolumn{1}{c|}{OJR} & \multicolumn{1}{c|}{Avg.} &
  EPM & ASI & \multicolumn{1}{c|}{HLD} & \multicolumn{1}{c|}{Avg.} &
  REC & SSR & \multicolumn{1}{c|}{CRR} & \multicolumn{1}{c|}{Avg.} &
  \multicolumn{1}{c}{} \\
  \midrule
  \multicolumn{18}{c}{\textbf{Human Evaluation}} \\ \midrule
  \multicolumn{1}{l|}{Human} & \multicolumn{1}{c|}{-} &
  93.9 & 92.5 & 94.8 & 92.7 & 91.0 & \multicolumn{1}{c|}{94.0} & \multicolumn{1}{c|}{93.2} &
  92.5 & 93.0 & \multicolumn{1}{c|}{91.4} & \multicolumn{1}{c|}{92.3} &
  95.4 & 89.6 & \multicolumn{1}{c|}{93.5} & \multicolumn{1}{c|}{92.8} & 92.8 \\
  \midrule
  \multicolumn{18}{c}{\textbf{Open-source Offline Models}} \\ \midrule
  \multicolumn{1}{l|}{MiniCPM-V4.5-8B~\citep{yu2025minicpm}} & \multicolumn{1}{c|}{0.5fps} &
  43.0 & 48.6 & 47.8 & 36.0 & 38.1 & \multicolumn{1}{c|}{41.6} & \multicolumn{1}{c|}{42.5} &
  33.2 & 55.7 & \multicolumn{1}{c|}{43.8} & \multicolumn{1}{c|}{44.2} &
  16.2 & 23.1 & \multicolumn{1}{c|}{5.4} & \multicolumn{1}{c|}{14.9} & 34.4 \\
  \multicolumn{1}{l|}{InternVL-3.5-8B~\citep{wang2025internvl3}} & \multicolumn{1}{c|}{0.5fps} &
  65.8 & 51.4 & 52.6 & 40.2 & 58.4 & \multicolumn{1}{c|}{53.8} & \multicolumn{1}{c|}{53.7} &
  46.0 & 52.7 & \multicolumn{1}{c|}{21.0} & \multicolumn{1}{c|}{39.9} &
  16.9 & 22.9 & \multicolumn{1}{c|}{4.0} & \multicolumn{1}{c|}{14.6} & 40.5 \\
  \multicolumn{1}{l|}{Qwen2-VL-8B~\citep{wang2024qwen2}} & \multicolumn{1}{c|}{0.5fps} &
  75.2 & 58.3 & 61.2 & 33.7 & 50.0 & \multicolumn{1}{c|}{59.8} & \multicolumn{1}{c|}{56.4} &
  48.2 & 44.3 & \multicolumn{1}{c|}{57.0} & \multicolumn{1}{c|}{49.8} &
  16.0 & 21.2 & \multicolumn{1}{c|}{0.8} & \multicolumn{1}{c|}{12.6} & 43.8 \\
  \multicolumn{1}{l|}{Qwen2.5-VL-8B~\citep{Qwen2.5-VL}} & \multicolumn{1}{c|}{0.5fps} &
  78.5 & 47.7 & 58.6 & 38.2 & 56.4 & \multicolumn{1}{c|}{58.2} & \multicolumn{1}{c|}{56.3} &
  41.6 & \textbf{56.8} & \multicolumn{1}{c|}{42.5} & \multicolumn{1}{c|}{46.9} &
  14.9 & 15.0 & \multicolumn{1}{c|}{8.0} & \multicolumn{1}{c|}{12.6} & 43.0 \\
  \multicolumn{1}{l|}{Qwen3-VL-8B~\citep{Qwen3-VL}} & \multicolumn{1}{c|}{0.5fps} &
  54.7 & 44.5 & 57.3 & 43.8 & 35.2 & \multicolumn{1}{c|}{56.0} & \multicolumn{1}{c|}{48.6} &
  34.2 & 50.3 & \multicolumn{1}{c|}{38.2} & \multicolumn{1}{c|}{40.9} &
  17.1 & 24.6 & \multicolumn{1}{c|}{5.6} & \multicolumn{1}{c|}{15.8} & 38.5 \\
  \multicolumn{1}{l|}{LLaVA-Video-7B~\citep{zhang2024llavavideo}} & \multicolumn{1}{c|}{0.5fps} &
  50.3 & 52.3 & 49.6 & 31.5 & 51.5 & \multicolumn{1}{c|}{41.3} & \multicolumn{1}{c|}{46.1} &
  38.4 & 51.4 & \multicolumn{1}{c|}{16.1} & \multicolumn{1}{c|}{35.3} &
  17.8 & 23.2 & \multicolumn{1}{c|}{4.9} & \multicolumn{1}{c|}{15.3} & 35.7 \\
  \multicolumn{1}{l|}{LLaVA-OneVision-7B~\citep{li2024llava}} & \multicolumn{1}{c|}{0.5fps} &
  47.3 & 50.5 & 63.8 & 46.1 & 51.5 & \multicolumn{1}{c|}{54.1} & \multicolumn{1}{c|}{52.2} &
  37.5 & 49.3 & \multicolumn{1}{c|}{\textbf{67.7}} & \multicolumn{1}{c|}{\textbf{51.5}} &
  13.0 & 20.2 & \multicolumn{1}{c|}{1.2} & \multicolumn{1}{c|}{11.5} & 41.9 \\
\multicolumn{1}{l|}{Video-LLaMA2-7B~\citep{damonlpsg2024videollama2}} & \multicolumn{1}{c|}{0.5fps} &
  20.8 & 28.4 & 17.7 & 15.7 & 29.7 & \multicolumn{1}{c|}{24.5} & \multicolumn{1}{c|}{22.8} &
  14.8 & 30.4 & \multicolumn{1}{c|}{7.5} & \multicolumn{1}{c|}{17.6} &
  13.6 & 21.6 & \multicolumn{1}{c|}{1.3} & \multicolumn{1}{c|}{12.2} & 18.9 \\
  \midrule
  \multicolumn{18}{c}{\textbf{Open-source Online Models}} \\ \midrule
  \multicolumn{1}{l|}{Flash-VStream-7B~\citep{zhang2024flashvstream}} & \multicolumn{1}{c|}{0.5fps} &
  24.2 & 29.4 & 28.5 & 33.7 & 25.7 & \multicolumn{1}{c|}{28.8} & \multicolumn{1}{c|}{28.4} &
  39.1 & 37.2 & \multicolumn{1}{c|}{5.9} & \multicolumn{1}{c|}{27.4} &
  17.4 & 13.5 & \multicolumn{1}{c|}{5.1} & \multicolumn{1}{c|}{12.0} & 24.0 \\
  \multicolumn{1}{l|}{Dispider-7B~\citep{qian2025dispider}} & \multicolumn{1}{c|}{0.5fps} &
  47.0 & 44.0 & 45.7 & 37.1 & 58.4 & \multicolumn{1}{c|}{48.4} & \multicolumn{1}{c|}{46.8} &
  46.5 & 51.4 & \multicolumn{1}{c|}{3.2} & \multicolumn{1}{c|}{33.7} &
  5.0 & 0.0 & \multicolumn{1}{c|}{12.4} & \multicolumn{1}{c|}{5.8} & 33.3 \\
  \multicolumn{1}{l|}{TimeChat-Online~\citep{yao2025timechatonline}} & \multicolumn{1}{c|}{0.5fps} &
  24.2 & 30.3 & 31.0 & 25.3 & 48.5 & \multicolumn{1}{c|}{36.4} & \multicolumn{1}{c|}{32.6} &
  37.7 & 41.9 & \multicolumn{1}{c|}{19.9} & \multicolumn{1}{c|}{33.2} &
  14.7 & 17.3 & \multicolumn{1}{c|}{3.3} & \multicolumn{1}{c|}{11.8} & 27.5 \\
  \midrule
  \multicolumn{18}{c}{\textbf{EvoStreaming Framework}} \\
  \midrule
  \rowcolor{SkyBlue!20}\multicolumn{1}{l|}{\MethodName-8B-MiniCPM-V} & \multicolumn{1}{c|}{0.5fps} &
  46.3 & 45.0 & 41.4 & 29.2 & 45.5 & \multicolumn{1}{c|}{42.4} & \multicolumn{1}{c|}{41.6} &
  41.1 & 46.0 & \multicolumn{1}{c|}{21.5} & \multicolumn{1}{c|}{36.2} &
  23.1 & \textbf{71.4} & \multicolumn{1}{c|}{23.3} & \multicolumn{1}{c|}{39.3} & 39.7 \\
\rowcolor{SkyBlue!20}\multicolumn{1}{l|}{\MethodName-8B-InternVL3.5} & \multicolumn{1}{c|}{0.5fps} &
  59.2 & 51.3 & 47.0 & 34.7 & 52.0 & \multicolumn{1}{c|}{48.1} & \multicolumn{1}{c|}{48.7} &
  43.5 & 49.3 & \multicolumn{1}{c|}{21.2} & \multicolumn{1}{c|}{38.0} &
  \textbf{26.1} & 62.9 & \multicolumn{1}{c|}{\textbf{29.1}} & \multicolumn{1}{c|}{\textbf{39.4}} & 43.7 \\
  \rowcolor{SkyBlue!20}\multicolumn{1}{l|}{\MethodName-8B-Qwen3} & \multicolumn{1}{c|}{0.5fps} &
  79.9 & 50.5 & 62.1 & 37.1 & 38.6 & \multicolumn{1}{c|}{60.6} & \multicolumn{1}{c|}{54.8} &
  45.1 & 41.2 & \multicolumn{1}{c|}{33.6} & \multicolumn{1}{c|}{40.0} &
  19.3 & {69.3} & \multicolumn{1}{c|}{24.2} & \multicolumn{1}{c|}{37.6} & 46.8 \\
  \rowcolor{SkyBlue!20}\multicolumn{1}{l|}{\MethodName-8B-Qwen2.5} & \multicolumn{1}{c|}{0.5fps} &
  78.9 & 58.3 & 63.4 & 51.1 & 56.9 & \multicolumn{1}{c|}{60.9} & \multicolumn{1}{c|}{61.6} &
  45.3 & 53.0 & \multicolumn{1}{c|}{35.2} & \multicolumn{1}{c|}{44.5} &
  18.9 & 38.7 & \multicolumn{1}{c|}{7.9} & \multicolumn{1}{c|}{21.8} & 47.4 \\
  \rowcolor{SkyBlue!20}\multicolumn{1}{l|}{\MethodName-8B-Qwen2} & \multicolumn{1}{c|}{0.5fps} &
  \textbf{82.9} & \textbf{67.9} & \textbf{75.9} & \textbf{53.4} & \textbf{62.4} & \multicolumn{1}{c|}{\textbf{69.0}} & \multicolumn{1}{c|}{\textbf{68.6}} &
  \textbf{51.5} & 55.4 & \multicolumn{1}{c|}{45.2} & \multicolumn{1}{c|}{50.7} &
  20.6 & 64.6 & \multicolumn{1}{c|}{6.7} & \multicolumn{1}{c|}{30.6} & \textbf{54.6} \\
  \bottomrule
  \end{tabular}
  }
  \vspace{-0.5em}
\end{table*}

\subsection{Setup}
\label{sec:setup}

\noindent\textbf{Models.}
We evaluate EvoStreaming on five VideoLLM backbones spanning three model families: Qwen2-VL, Qwen2.5-VL, and Qwen3-VL~\citep{wang2024qwen2,Qwen2.5-VL,Qwen3-VL}, InternVL-3.5~\citep{wang2025internvl3}, and MiniCPM-V4.5~\citep{yu2025minicpm}.
These backbones cover different visual encoders, token compression strategies, and pre-training data mixtures, so that consistent gains can be attributed to the streaming adaptation recipe rather than backbone-specific artifacts.
Each backbone is adapted with the same $1{,}000$ self-evolved trajectories and identical hyperparameters; no per-model tuning is applied.
Baselines include the vanilla (unadapted) backbones, additional offline VideoLLMs (LLaVA-OneVision-7B~\citep{li2024llava}, Video-LLaMA2-7B~\citep{damonlpsg2024videollama2}), and dedicated streaming models (Flash-VStream-7B~\citep{zhang2024flashvstream}, Dispider-7B~\citep{qian2025dispider}, TimeChat-Online~\citep{yao2025timechatonline}).

\noindent\textbf{Benchmarks.}
For streaming evaluation we use the task definitions and QA data from OVO-Bench~\citep{li2025ovobench} under the RealStreamEval protocol (Section~\ref{sec:realestreameval}): frames arrive at 0.5\,fps and the model decides at each turn whether to respond or remain silent.
Because RealStreamEval adds a verbosity penalty and removes the chunked-input assumption, scores are not directly comparable to standard OVO-Bench numbers.
For offline evaluation we report VideoMME~\citep{fu2025video}, LVBench~\citep{wang2024lvbench}, LongVideoBench~\citep{wu2024longvideobench}, EgoSchema~\citep{mangalam2024egoschema}, and MLVU~\citep{zhou2024mlvu} to verify that streaming adaptation does not degrade general video understanding.

\noindent\textbf{Implementation.}
All backbones are fine-tuned with LoRA (rank 8, applied to all linear layers) using ms-swift~\citep{zhao2024swiftascalablelightweightinfrastructure}.
Visual tokens per frame are capped at 128 during training for efficiency; at inference we evaluate both 128 and 768 tokens (Table~\ref{tab:token_ablation}).
The LLM judge for RealStreamEval is Qwen3-VL-235B-A22B-Instruct~\citep{Qwen3-VL}, served via vLLM~\citep{kwon2023efficient}.
Offline benchmarks are evaluated with lmms-eval~\citep{zhang2024lmmsevalrealitycheckevaluation} under default settings.
Full hyperparameters are listed in Appendix Table~\ref{tab:hyperparameters}.

\begin{figure*}[!t]
    \centering
    \begin{subfigure}[t]{0.49\textwidth}
        \centering
        \includegraphics[width=\linewidth]{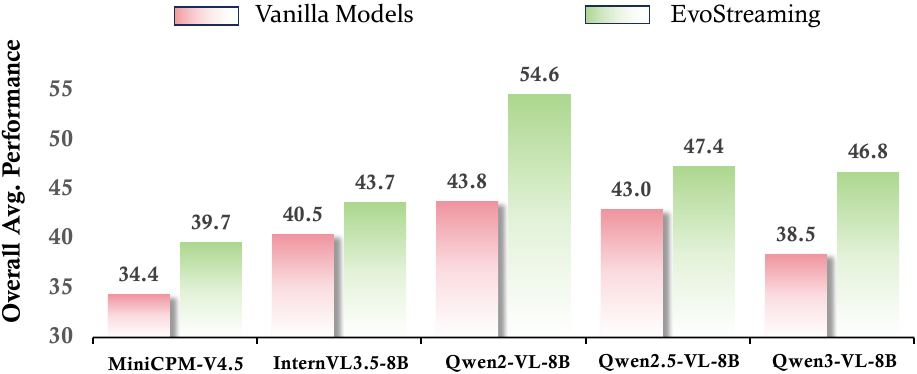}
        \caption{Overall OVO-Bench comparison.}
        \label{fig:vanilla_vs_evostreaming_overall}
    \end{subfigure}
    \hfill
    \begin{subfigure}[t]{0.49\textwidth}
        \centering
        \includegraphics[width=\linewidth]{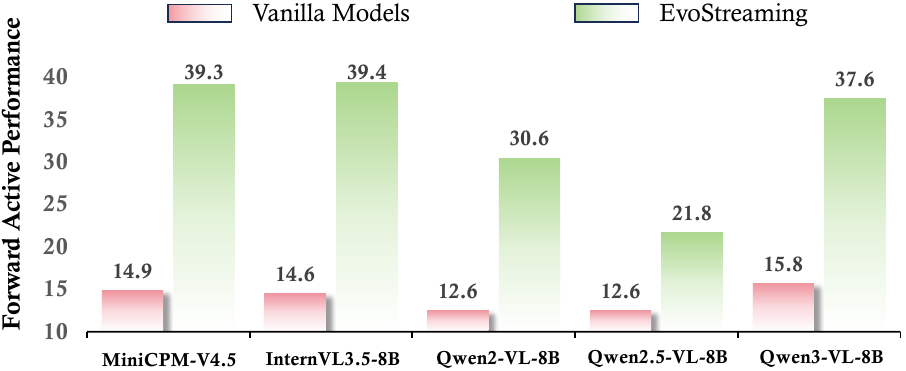}
        \caption{Forward Active Responding comparison.}
        \label{fig:Forward_Active}
    \end{subfigure}
    \caption{\textbf{Streaming performance analysis.}
    Left: EvoStreaming improves overall OVO-Bench performance across base models.
    Right: EvoStreaming improves Forward Active Responding, i.e., deciding when proactive responses are needed.}
    \label{fig:vanilla_vs_evostreaming}
\end{figure*}

\begin{table*}[!t]
\centering
\setlength{\tabcolsep}{2.5pt}
\renewcommand{\arraystretch}{1.0}
\caption{\textbf{Results on offline video benchmarks.} We report performance on VideoMME, LVBench, LongVideoBench, EgoSchema, and MLVU under the vanilla offline evaluation protocol.}
\label{tab:offline}
\resizebox{0.97\textwidth}{!}{%
\footnotesize
\begin{tabular}{@{}l|cccc|c|c|c|c|c@{}}
\toprule
\multirow{2}{*}{Model} & \multicolumn{4}{c|}{VideoMME} & \multirow{2}{*}{LVBench} & \multirow{2}{*}{LongVideoBench} & \multirow{2}{*}{EgoSchema} & \multirow{2}{*}{MLVU} & \multirow{2}{*}{Avg.} \\
& Overall & Short & Medium & Long &  &  &  &  &  \\ \midrule
\multicolumn{10}{c}{\textbf{Proprietary Models}} \\ \midrule
Gemini-1.5-pro~\citep{team2024gemini} & 75.0 & 81.7 & 74.3 & 67.4 & 33.1 & 64.0 & 71.2 & - & - \\
GPT-4o~\citep{hurst2024gpt} & 71.9 & 80.0 & 70.3 & 65.3 & 48.9 & 66.7 & 72.2 & 64.6 & 64.9 \\ \midrule
\multicolumn{10}{c}{\textbf{Open-source Offline Models}} \\ \midrule
LLaVA-OneVision-7B~\citep{li2024llava} & 58.2 & 69.1 & 53.3 & 46.7 & - & 56.3 & 59.8 & 64.7 & - \\
LLaVA-Video-7B~\citep{zhang2024llavavideo} & 63.3 & - & - & - & 41.5 & 58.2 & 57.3 & 70.8 & 58.2 \\
InternVL-V2-8B~\citep{chen2024far} & 56.3 & - & - & - & - & 54.6 & - & 64.0 & - \\
InternVL-V2.5-8B~\citep{chen2024expanding} & 64.2 & - & - & - & 43.2 & 60.0 & 51.5 & 68.9 & 57.6 \\

TimeChat~\citep{Ren2023TimeChat} & 30.2 & - & - & - & 22.3 & - & 33.0 & 30.9 & - \\
LLaVA-Next-Video-7B~\citep{zhang2024llavanextvideo} & 46.6 & - & - & - & - & 43.5 & 43.9 & - & - \\
LongVU-7B~\citep{shen2024longvu} & 60.9 & 64.7 & 58.2 & 59.5 & -  & - & 67.6 & 65.4& - \\
VideoChat2-7B~\citep{li2024mvbench} & 43.8 & 52.8 & 39.4 & 39.2 & - & 39.3 & 54.4 & 47.9 & - \\
LongVA-7B~\citep{zhang2024long} & 52.6 &61.1& 50.4 & 46.2 & - &- & 42.5 & 56.3 & - \\
Kangaroo-7B~\citep{liu2024kangaroo}&56.0 	&66.1 	&55.3 	&46.7 	&38.3 		&54.2 	&62.7 &	61.0 & 54.4\\
ShareGPT4-video~\citep{chen2024sharegpt4video} & 39.9 & 48.3 & 36.3 & 35.0 & - & 39.7 & - & 46.4 & - \\
Video-LLaMA2-7B~\citep{damonlpsg2024videollama2} & 47.9 & 56.0 & 45.4 & 42.0 & - & - & 51.7 & 48.5 & - \\
Video-LLaMA3-7B~\citep{zhang2025videollama} & 66.2 & 80.1 & 63.7 & 54.9 & 45.3 & 59.8 & 63.3 & 73.0 & 61.5 \\
\midrule
\multicolumn{10}{c}{\textbf{Open-source Online Models}} \\ \midrule
Flash-VStream-7B~\citep{zhang2024flashvstream} & 61.2 & 72.0 & 61.1 & 50.3 & 42.0 & - & 38.1 & 50.2 & - \\
Dispider-7B~\citep{qian2025dispider} & 57.2 & - & - & - & - & -& 55.6 & 61.7 & -  \\
TimeChat-Online-7B~\citep{yao2025timechatonline} & 53.2 & - & - & 41.7 & - & 57.1 & 61.9 & 62.9 & - \\
VideoLLM-online-8B~\citep{chen2024videollm} & 52.8 & - & - & 44.9 & 24.0 & - & 32.8 & 35.2 & - \\
\midrule
\multicolumn{10}{c}{\textbf{EvoStreaming Framework}} \\ \midrule

InternVL-V3.5-8B~\citep{wang2025internvl3} & 64.7 & 77.4 & 63.1 & 53.4 & 42.5 & 61.0 & 60.0 & 66.2 & 58.9 \\
\rowcolor{SkyBlue!20}
EvoStreaming-8B-InternVL-V3.5 & 63.5 & 76.7 & 62.0 & 51.9 & 42.2 & 59.2 & 58.0 & 66.0 & 57.8 \\

\midrule
Qwen2.5-VL-7B~\citep{Qwen2.5-VL} & 62.0 & 73.0 & 60.9 & 52.1 & 40.2  & 58.3 & 63.2 & 58.7 & 56.5 \\
\rowcolor{SkyBlue!20}
EvoStreaming-8B-Qwen2.5 & 61.9 &73.1 	& 61.0	& 51.4 	& 39.8 	 &	58.1 	& 58.9  & 	60.1  & 55.8 \\ \midrule

Qwen2-VL-7B~\citep{wang2024qwen2} & 57.9 & 70.2  & 54.6 & 48.8 & 39.1 & 55.5 & 62.2 & 60.1  & 55.0 \\
\rowcolor{SkyBlue!20}
EvoStreaming-8B-Qwen2 & 57.9 & 69.4  &	55.2  &	48.9	& 38.8 &	53.4 & 57.9  & 60.2 & 53.6 \\

\bottomrule
\end{tabular}%
}
\end{table*}
\subsection{Main Results}
\label{sec:main_results}


\noindent\textbf{Streaming evaluation.}
As shown in Table~\ref{tab:main_exp} and Figure~\ref{fig:vanilla_vs_evostreaming}, we conducted comprehensive streaming experiments on five VideoLLM backbones.
Compared with representative online baselines, EvoStreaming substantially outperforms module-augmented streaming adaptation methods: EvoStreaming-Qwen2 reaches \textbf{54.6\%}, exceeding Dispider (33.3\%) by $+$21.3 average points. This margin suggests that directly learning interaction timing can be more effective than attaching auxiliary streaming controllers to offline backbones. It also reinforces that strong offline VideoLLMs can be transferred to streaming settings with relatively lightweight adaptation, whereas additional streaming-specific modules may increase adaptation complexity and reduce generality.
Beyond outperforming baselines, the gains are also strikingly data-efficient. With only $1{,}000$ self-generated trajectories and no architectural changes, EvoStreaming improves every adapted backbone by $+$10.8 (Qwen2), $+$4.4 (Qwen2.5), $+$8.3 (Qwen3), $+$3.2 (InternVL-3.5), and $+$5.3 (MiniCPM-V4.5) over vanilla counterparts, indicating that high-value timing supervision can substitute for large-scale streaming annotation.
These improvements are not uniformly distributed but concentrate on the most streaming-critical abilities. For Qwen2, Real-Time Visual Perception rises from 56.4\% to 68.6\%, and Forward Active Responding from 12.6\% to 30.6\%, indicating better evidence grounding and more reliable response-timing control under sequential observations.

\noindent\textbf{Offline preservation.}
Although our primary focus is streaming video understanding, an equally important question is whether adaptation preserves the backbone's original offline competence.
Under a strictly streaming-only setting with just $1{,}000$ trajectories and no additional offline-video training samples, the offline performance shift remains small across models.
Averaged over five offline benchmarks, Qwen2.5 changes from 56.5\% to 55.8\%, Qwen2 from 55.0\% to 53.6\%, and InternVL-3.5 by 1.1 points, indicating only mild degradation after adaptation.
This pattern is further supported by VideoMME, where Qwen2 stays unchanged at 57.9\% before and after streaming fine-tuning.
Beyond the per-backbone view, our adapted models also remain highly competitive against representative online video baselines: on VideoMME overall, EvoStreaming-InternVL-3.5 reaches 63.5\% and EvoStreaming-Qwen2.5 reaches 61.9\%, both clearly ahead of Flash-VStream (61.2\%), Dispider (57.2\%), TimeChat-Online (53.2\%), and VideoLLM-Online (52.8\%).
Taken together, these results suggest that lightweight tuning on a small self-generated streaming set can deliver substantial online gains while largely preserving the backbone's general offline video understanding ability, and even leaves our streaming-adapted models highly competitive on standard offline benchmarks.

\section{Analysis}
\label{sec:analysis}

\begin{figure*}[!t]
    \centering
    \includegraphics[width=\linewidth, height=0.36\linewidth]{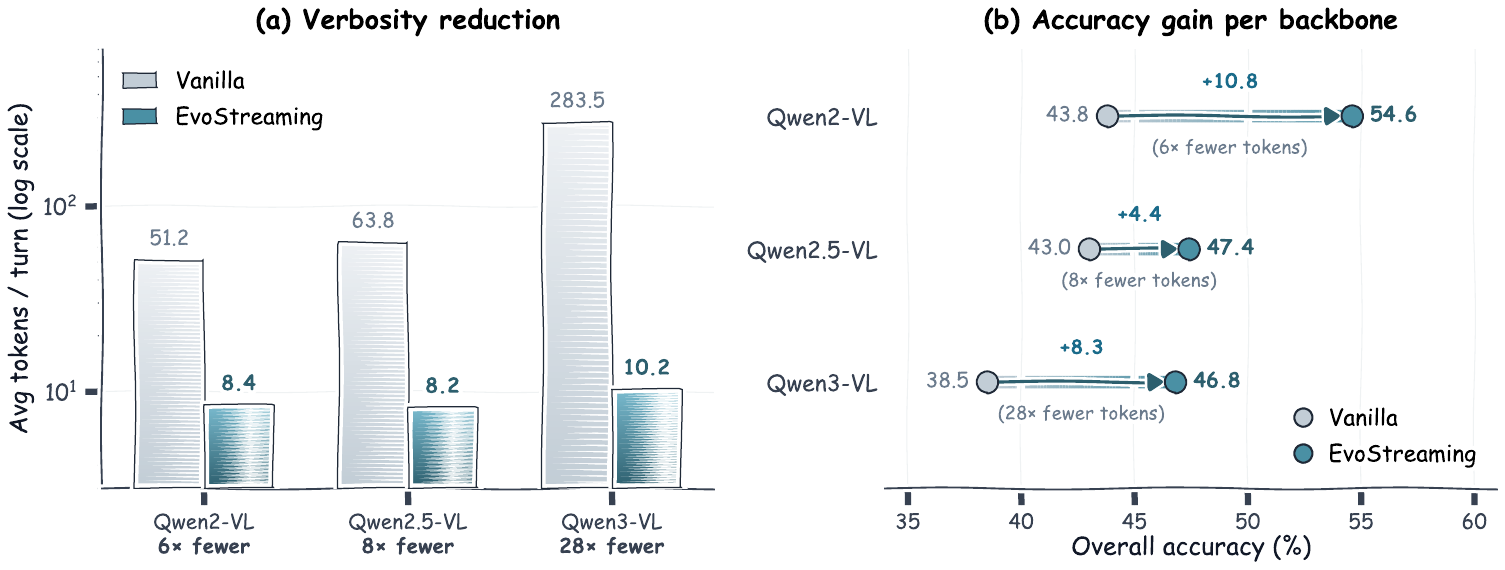}
    \vspace{-1.5em}
    \caption{\textbf{EvoStreaming reduces verbosity while improving accuracy.} (a) Average tokens per turn for three Qwen-family backbones, with $6\times$--$28\times$ fewer tokens. (b) Accuracy gains per backbone from vanilla to EvoStreaming. Detailed numbers are in Appendix Table~\ref{tab:per_token_score}.}
    \label{fig:per_token}
    \vspace{-1em}
\end{figure*}

\noindent \textbf{Why Does Self-Evolution Succeed with Minimal Data?}
The sample efficiency of EvoStreaming rests on two structural properties of the adaptation problem. First, streaming video understanding decomposes naturally into visual comprehension and interaction timing, and the base model has already acquired the former during offline pre-training. What remains to be learned is a comparatively low-dimensional timing policy that maps visual-linguistic context to a binary \textsc{Silent}/respond decision. Because LoRA adaptation freezes the pretrained encoder and only introduces a low-rank perturbation, the effective hypothesis class is a small timing head rather than the full model, so the sample budget scales with the timing-head capacity and not with the encoder parameter count. Second, the self-generated labels are not adversarial noise: the binary relevance judgement in Stage~3 is class-conditionally noisy at a rate bounded by the encoder's own segment-level relevance error $\epsilon_V$, and when the base model is a competent VideoLLM ($\epsilon_V \ll 1/2$) the noise-inflation factor stays close to $1$. The two effects together predict that a few hundred to a thousand trajectories should suffice, which matches the consistent gains we observe across all five backbones. Please refer to Appendix~\ref{sec:theory} for the full derivation.

\noindent \textbf{Training Efficiency and Resolution Robustness.}
A timing policy that can be learned from coarse visual features would make streaming adaptation practical on resource-constrained hardware. We therefore train at only $128$ tokens per frame and evaluate whether the policy survives a switch to $768$-token inference. The overall RealStreamEval score shifts by at most $1.4$ points across the three Qwen backbones (Table~\ref{tab:token_ablation}), confirming that timing decisions are largely resolution-agnostic. The resolution-sensitive gains concentrate on Real-Time Visual Perception, where EvoStreaming-Qwen2 rises from $62.1$ to $68.6$ thanks to improved OCR and small-object recognition, and on Forward Active Responding, where Qwen3 benefits most from $768$ tokens ($+11.3$, from $26.3$ to $37.6$) because finer visual detail helps ground causal reasoning in CRR. The timing head itself needs no retraining for higher resolution; the encoder simply sees more pixels.

\noindent \textbf{Why Stronger Offline Models Do Not Always Score Higher?}
Under RealStreamEval, the Qwen-family ordering is reversed relative to standard offline benchmarks: Qwen2-VL (43.8\%) outperforms Qwen2.5-VL (43.0\%) and Qwen3-VL (38.5\%) (Table~\ref{tab:main_exp}). To diagnose the reversal, we compute a \emph{Per-Token Score} $\eta = \text{Overall}/\text{Avg.\ tokens-per-turn}$, which captures how much benchmark credit a model earns per generated token and exposes the streaming-specific cost of verbosity that raw accuracy hides (Appendix~\ref{sec:verbosity_analysis}). As shown in Figure~\ref{fig:per_token}, newer Qwen versions emit substantially more tokens per turn in vanilla mode (283 for Qwen3-VL vs. 51 for Qwen2-VL), which inflates the answer rate and triggers the verbosity penalty. EvoStreaming closes this gap by bringing all three backbones to ${\sim}10$ tokens per turn. The effect is most visible on Qwen3-VL: starting from the lowest baseline of 38.5\% (a direct consequence of its 283-token vanilla response rate), it recovers +8.3 points, confirming that the Qwen-family reversal was driven by verbosity rather than by visual understanding.

\section{Conclusion}
\label{sec:conclusion}


In this work, we study how offline VideoLLMs can be adapted and evaluated under streaming interaction constraints. We introduce \textbf{RealStreamEval}, a frame-level multi-turn evaluation protocol that jointly considers answer correctness, response timing, and unnecessary responses. We further propose \textbf{EvoStreaming}, a self-evolved streaming adaptation framework that uses $1{,}000$ self-generated streaming trajectories ($139\times$ less than the leading streaming video model) and no architectural changes to teach when a model should respond or remain silent. Across five open VideoLLM backbones, we improve the overall RealStreamEval score by up to $10.8$ while largely preserving offline video performance. More broadly, our results suggest that lightweight interaction-timing supervision can turn strong offline VideoLLMs into practical streaming assistants without relying on expensive redesign or large-scale streaming annotation.


\bibliographystyle{colm2026_conference}
\bibliography{main}

\clearpage
\appendix

\let\addcontentsline\originaladdcontentsline
\startcontents[appendix]
\printcontents[appendix]{ }{0}{\section*{Appendix}}

\section{Implementation Details}
\label{sec:implementation_appendix}

We provide additional implementation details for both the self-evolution data construction (Algorithm~\ref{alg:evo}) and the fine-tuning of EvoStreaming-adapted models. All experiments use ms-swift~\citep{zhao2024swiftascalablelightweightinfrastructure} as the training framework. Fine-tuning is performed with LoRA~\citep{hu2022lora} to keep the parameter footprint small and to match the assumptions used in the sample-efficiency analysis in Appendix~\ref{sec:theory}. We use the same training recipe across all five base backbones (Qwen2-VL, Qwen2.5-VL, Qwen3-VL, InternVL-3.5, MiniCPM-V4.5) without per-model tuning, so the hyperparameters listed in Table~\ref{tab:hyperparameters} apply uniformly unless otherwise noted.

\begin{table*}[!t]
  \centering
  \setlength{\tabcolsep}{8pt}
  \renewcommand{\arraystretch}{1.18}
  \caption{\textbf{Implementation hyperparameters used for EvoStreaming.} Settings are shared across all five base backbones unless otherwise noted.}
  \label{tab:hyperparameters}
  \resizebox{0.72\textwidth}{!}{%
  \begin{tabular}{@{}ll@{}}
  \toprule
  \textbf{Hyperparameter} & \textbf{Value} \\
  \midrule
  \multicolumn{2}{l}{\textit{Self-evolution data construction (Algorithm~\ref{alg:evo})}} \\
  \midrule
  Source video pool $\mathbb{V}$                                   & $11{,}043$ (TimeChat-Online-139K) \\
  Number of self-generated trajectories $|\mathcal{D}|$            & $1{,}087$ \\
  Outer iterations $I$                                             & $1$, $2$, or $3$ \\
  Number of questions per video $K$                                & $2.7$ on average \\
  Frame sampling rate                                              & $0.5$ fps \\
  Visual tokens per frame (training)                               & $128$ \\
  Task-type categories                                             & $5$ (IV, MD, TA, AM, DED) \\
  \midrule
  \multicolumn{2}{l}{\textit{Fine-tuning configuration}} \\
  \midrule
  Training framework                                               & ms-swift~\citep{zhao2024swiftascalablelightweightinfrastructure} \\
  Adaptation method                                                & LoRA~\citep{hu2022lora} \\
  LoRA rank $r$                                                    & $8$ \\
  LoRA scaling $\alpha$                                            & $32$ \\
  LoRA target modules                                              & all linear layers \\
  LoRA dropout                                                     & $0.05$ \\
  Learning rate                                                    & $1\times10^{-4}$ \\
  Learning-rate schedule                                           & cosine \\
  Warmup ratio                                                     & $0.05$ \\
  Optimizer                                                        & AdamW \\
  Weight decay                                                     & $0.1$ \\
  Global batch size                                                & $8$ \\
  Per-device batch size                                            & $1$ \\
  Gradient accumulation steps                                      & $1$ \\
  Number of epochs (per outer iteration)                           & $1$ \\
  Mixed precision                                                  & bf16 \\
  Gradient clipping                                                & $1.0$ \\
  Random seed                                                      & $42$ \\
  \midrule
  \multicolumn{2}{l}{\textit{Hardware and runtime}} \\
  \midrule
  GPU type                                                         & NVIDIA H200 141\,GB \\
  Number of GPUs                                                   & $8$ \\
  \midrule
  \multicolumn{2}{l}{\textit{Inference-time settings (RealStreamEval)}} \\
  \midrule
  Visual tokens per frame (inference)                              & $128$ / $768$ \\
  LLM judge model                                                  & Qwen3-VL-235B-A22B-Instruct~\citep{Qwen3-VL} \\
  \bottomrule
  \end{tabular}%
  }
\end{table*}

\begin{figure}[!t]
    \centering
    \includegraphics[width=0.88\textwidth]{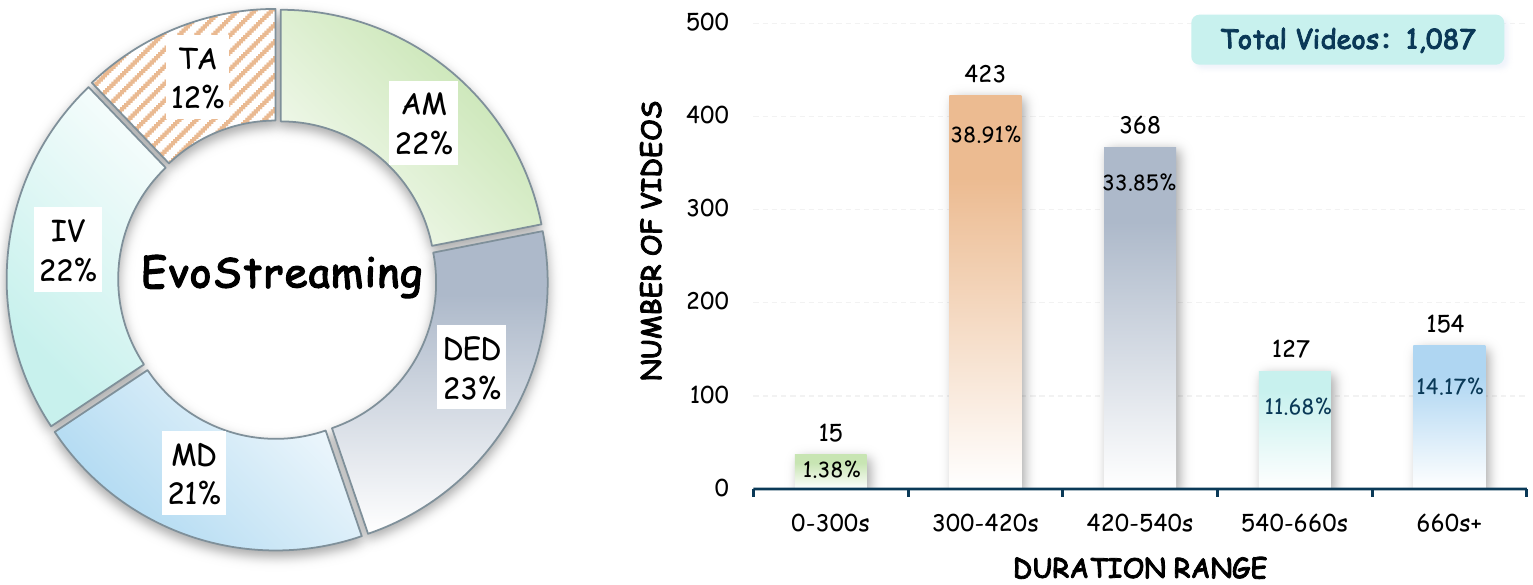}
    \caption{Dataset distribution overview. Left: task distribution; Right: video duration distribution.}
    \label{fig:data_distribution}
\end{figure}

\subsection{Task-Type Taxonomy for Self-Evolved Trajectories}
\label{sec:task_taxonomy}

The five task categories used in Stage~1 of Algorithm~\ref{alg:evo} are defined to cover the spectrum of streaming behaviors that a real-time assistant must support, ranging from instantaneous reactions to sustained patient monitoring. They serve as generation-side priors that diversify the self-synthesized $\mathcal{D}$, not as evaluation labels. The categories and their intended semantics are summarized in Table~\ref{tab:task_taxonomy}, together with their loose correspondence to the three OVO-Bench task modes used at evaluation time. The mapping is one-to-many on both sides: a single self-generated category can produce trajectories that probe several OVO-Bench subtasks, and a given subtask can be hit by trajectories from more than one category. This loose mapping is intentional: the taxonomy is designed to elicit the timing patterns we want the model to internalize (instant response, retrospective recall, patient anticipation, etc.), rather than to one-to-one match an evaluator's subtask list.

\begin{table*}[!t]
  \centering
  \setlength{\tabcolsep}{6pt}
  \renewcommand{\arraystretch}{1.15}
  \caption{\textbf{Self-evolved task-type taxonomy.} The five categories used by Stage~1 of Algorithm~\ref{alg:evo} cover the streaming-interaction spectrum from instantaneous reaction to patient observation. The right column gives the typical OVO-Bench mode that each category exercises; the mapping is loose by design.}
  \label{tab:task_taxonomy}
  \resizebox{\textwidth}{!}{%
  \begin{tabular}{@{}llp{0.5\textwidth}l@{}}
  \toprule
  \textbf{Category} & \textbf{Code} & \textbf{Intended streaming behavior} & \textbf{Typical OVO-Bench mode} \\
  \midrule
  Immediate Visual & IV & Respond as soon as evidence is visible in the current segment; describe what is happening right now (e.g.\ recognize an object, read a sign, identify an action). & Real-Time Visual Perception \\
  Memory-Dependent & MD & Recall a past event from the already-observed history when asked retroactively (e.g.\ ``what color was the car that just passed?''). & Backward Tracing \\
  Temporal Aggregation & TA & Aggregate evidence across multiple segments before producing an answer (e.g.\ counting repeated actions, summarizing a sequence of steps). & Backward Tracing / Forward Active Responding \\
  Anticipatory Monitoring & AM & Stay silent until a triggering visual cue appears, then respond promptly (e.g.\ alert when a specific event begins, raise a clue-based causal answer). & Forward Active Responding \\
  Dynamic Event Description & DED & Continuously describe an unfolding event with appropriately spaced updates (e.g.\ narrate a procedure as it progresses), exercising both \textsc{Silent} restraint between updates and timely responses at salient transitions. & Spans all three modes \\
  \bottomrule
  \end{tabular}}
\end{table*}

\section{More Results}
\label{sec:more_results}


\subsection{Performance on the Original OVO-Bench Protocol}
\label{sec:original_ovobench}

Although the original OVO-Bench~\citep{li2025ovobench} adopts an offline chunked evaluation protocol that is not fully aligned with the streaming objective EvoStreaming is trained for, we additionally report EvoStreaming's performance under this mismatched setting to verify that streaming adaptation does not degrade the base model's general offline video understanding. Table~\ref{tab:original_ovo} compares two base models with their EvoStreaming-adapted counterparts under default OVO-Bench inference. The EvoStreaming-adapted variant slightly outperforms its base model on the Overall Avg.\ score in both cases: $+0.6$ points on Qwen2-VL ($50.35\!\to\!50.94$) and $+0.8$ points on MiniCPM-V4.5 ($60.98\!\to\!61.77$). The gains are concentrated on Backward Tracing, where both adapted variants improve by $2.3$ and $3.9$ points, while Real-Time Visual Perception scores remain essentially unchanged and Forward Active Responding shows small mixed changes. The Backward Tracing improvement is consistent with the temporal-tracking habits induced by streaming training: EvoStreaming requires the model to silently follow visual events before deciding when to respond, which appears to transfer naturally to retrospective tracing tasks. The streaming objective EvoStreaming is trained against therefore does not degrade, and slightly enhances, the base model's offline OVO-Bench performance.

\begin{table*}[!t]
  \centering
  \setlength{\tabcolsep}{2.5pt}
  \renewcommand{\arraystretch}{1.2}
  \caption{\textbf{Results on the original OVO-Bench (offline chunked evaluation).} For two backbones, we compare the base model (top of each pair) with its EvoStreaming-adapted variant (highlighted). EvoStreaming-adapted variants slightly outperform their base models on the Overall Avg.\ score ($+0.6$ on Qwen2-VL and $+0.8$ on MiniCPM-V4.5), with the largest gains concentrated on Backward Tracing ($+2.3$ and $+3.9$ points).}
  \label{tab:original_ovo}
  \resizebox{\textwidth}{!}{%
  \begin{tabular}{@{}lccccccccccccccc@{}}
  \toprule
  \multirow{2}{*}{Model} &
  \multicolumn{6}{c}{Real-Time Visual Perception} &
  \multicolumn{4}{c}{Backward Tracing} &
  \multicolumn{4}{c}{Forward Active Responding} &
  Overall \\
  \cmidrule(lr){2-7}\cmidrule(lr){8-11}\cmidrule(lr){12-15}
  & ACR & ATR & STU & FPD & OJR & Avg. & EPM & ASI & HLD & Avg. & REC & SSR & CRR & Avg. & Avg. \\
  \midrule
  Qwen2-VL-8B & 49.54 & 55.17 & 46.63 & 68.32 & 55.43 & 55.69 & 48.48 & 55.41 & 35.48 & 46.46 & 31.23 & 66.30 & 49.17 & 48.90 & 50.35 \\
  \rowcolor{SkyBlue!20}\MethodName-8B-Qwen2 & 46.79 & 58.62 & 50.00 & 66.34 & 54.89 & 56.29 & 48.82 & 55.41 & 41.94 & 48.72 & 33.38 & 60.10 & 50.00 & 47.83 & 50.94 \\
  \midrule
  MiniCPM-V4.5-8B & 58.72 & 74.14 & 55.62 & 69.31 & 63.59 & 66.65 & 59.93 & 62.84 & 57.53 & 60.10 & 36.39 & 66.77 & 65.42 & 56.19 & 60.98 \\
  \rowcolor{SkyBlue!20}\MethodName-8B-MiniCPM-V & 56.88 & 75.00 & 54.49 & 69.31 & 58.70 & 65.48 & 62.29 & 66.22 & 63.44 & 63.98 & 35.96 & 62.80 & 68.75 & 55.84 & 61.77 \\
  \bottomrule
  \end{tabular}%
  }
\end{table*}

\section{Discussion}

\subsection{A Sample-Efficiency Analysis of Self-Evolved Streaming Adaptation}
\label{sec:theory}

This section formalizes the two beliefs underlying Algorithm~\ref{alg:evo}: (i)~freezing\,/\,low-rank-restricting the pretrained visual encoder shrinks the effective hypothesis class to a small timing head, so few labelled trajectories suffice; (ii)~labels generated by the model itself are not adversarial: they are class-conditional noisy versions of the true labels with a noise rate that is bounded by the encoder's relevance error, so a standard noise-corrected ERM converges to the same optimum as clean supervision. Combining the two yields a quantitative bound that is consistent with our experimental observations and predicts when self-evolution should fail.

\paragraph{Setting and assumptions.}
Let $h_t = \phi_{\alpha}(\mathcal{H}_t) \in \mathcal{Z}$ denote the visual--linguistic context produced by the encoder of a base VideoLLM with parameters $\alpha$, and let the streaming policy factor as $\pi_\theta(\mathcal{H}_t) = g_\beta(h_t)$, where $g_\beta:\mathcal{Z}\to\mathcal{A}$ is a timing head. Offline pre-training yields $\hat\alpha$. EvoStreaming adapts the model with LoRA~\citep{hu2022lora}, which constrains the update to a low-rank perturbation $\Delta\alpha = BA$ with $\mathrm{rank}(BA)\le r$. We treat $\hat\alpha$ as effectively frozen for the purpose of analysis: the LoRA update modifies a small number of additional parameters that we absorb into the timing-head class. Concretely, let $\mathcal{B}$ be the (LoRA-augmented) timing-head hypothesis class with covering number $\mathcal{N}(\mathcal{B}, \cdot)$, and let $\ell:\mathcal{A}\times\mathcal{A}\to[0,B]$ be a bounded surrogate loss. Streaming adaptation minimizes the empirical risk
\begin{equation}
\widehat{R}_n(\beta) \;=\; \frac{1}{n}\sum_{i=1}^n \ell\bigl(g_\beta(\phi_{\hat\alpha}(\mathcal{H}_i)),\, a^{\star}_i\bigr).
\end{equation}

\paragraph{Proposition 1 (Decoupled sample complexity).}\label{prop:decoupled}
Under the LoRA-restricted setting above, with probability at least $1-\delta$ over $n$ i.i.d.\ samples,
\begin{equation}
\label{eq:prop1}
R(\hat\beta) - \inf_{\beta\in\mathcal{B}} R(\beta) \;\le\; 4B\sqrt{\frac{\log\mathcal{N}\!\bigl(\mathcal{B},\,n^{-1/2}\bigr) + \log(2/\delta)}{n}}.
\end{equation}
The corresponding bound for end-to-end joint training over $(\alpha,\beta)\in\mathcal{A}\!\times\!\mathcal{B}$ replaces $\log\mathcal{N}(\mathcal{B},\cdot)$ with $\log\mathcal{N}(\mathcal{A}\!\times\!\mathcal{B},\cdot)$.

\textit{Proof sketch.} Standard uniform convergence over $\mathcal{B}$ via Massart's lemma applied to the fixed feature map $\phi_{\hat\alpha}$, combined with a McDiarmid-type concentration on the empirical Rademacher complexity~\citep{bartlett2002rademacher}. The joint version follows by replacing $\mathcal{B}$ with the product class $\mathcal{A}\!\times\!\mathcal{B}$.\hfill$\square$

\paragraph{Implication.} For an 8B-parameter video transformer, $\log\mathcal{N}(\mathcal{A},\cdot)$ grows essentially linearly in the parameter count and is many orders of magnitude larger than $\log\mathcal{N}(\mathcal{B},\cdot)$, since (a) the LoRA rank is small ($r\!\le\!64$ in our experiments) and (b) the timing head distinguishes only $\textsc{Silent}$ from a triggered response on a short context window. Eq.~\eqref{eq:prop1} therefore predicts that the sample budget needed to reach a target excess risk is several orders of magnitude smaller for LoRA-restricted adaptation than for end-to-end training, which is one quantitative explanation for why $1{,}000$ trajectories already activate the timing policy across all five backbones we study.

\paragraph{Proposition 2 (Robustness to self-generated supervision).}\label{prop:noise}
Let the visual encoder produce segment-level relevance labels $\hat a_{t,k}\in\{\textsc{I},\textsc{R}\}$ in Stage~3 of Algorithm~\ref{alg:evo} with class-conditional noise rates
\begin{equation}
\rho_- = \Pr(\hat a = \textsc{R}\mid a^{\star} = \textsc{I}), \qquad \rho_+ = \Pr(\hat a = \textsc{I}\mid a^{\star} = \textsc{R}),
\end{equation}
both at most $\epsilon_V < 1/2$. Define the noise-corrected loss
\begin{equation}
\label{eq:correctedloss}
\tilde\ell\bigl(g(z),\hat a\bigr) \;=\; \frac{(1-\rho_{1-\hat a})\,\ell\bigl(g(z),\hat a\bigr) \,-\, \rho_{\hat a}\,\ell\bigl(g(z),\, 1-\hat a\bigr)}{1 - \rho_- - \rho_+}.
\end{equation}
Then the noise-corrected ERM is unbiased,
\begin{equation}
\mathbb{E}_{\hat a \mid a^{\star}}\!\bigl[\tilde\ell\bigl(g(z),\hat a\bigr)\bigr] \;=\; \ell\bigl(g(z),\,a^{\star}\bigr),
\end{equation}
so it converges to the same population minimizer as clean supervision, and its empirical Rademacher complexity inflates by at most a factor of $1/(1-2\epsilon_V)$. Hence $n$ self-generated labels statistically match $(1-2\epsilon_V)^2 n$ clean labels.

\textit{Proof sketch.} Direct verification of the unbiasedness identity by case analysis on $a^{\star}\in\{\textsc{I},\textsc{R}\}$, followed by a symmetrization argument bounding the Rademacher complexity of $\tilde\ell\circ\mathcal{B}$ in terms of that of $\ell\circ\mathcal{B}$. The result is the binary-noise specialization of Theorem~3 in~\citep{natarajan2013learning}.\hfill$\square$

\paragraph{Combined bound.} Substituting the noise-corrected loss of Proposition~\ref{prop:noise} into the uniform-convergence bound of Proposition~\ref{prop:decoupled} yields
\begin{equation}
\label{eq:combined}
n \;=\; \widetilde{\mathcal{O}}\!\left(\frac{\log\mathcal{N}\!\bigl(\mathcal{B},\,n^{-1/2}\bigr)}{(1-2\epsilon_V)^2 \,\epsilon^2}\right)
\end{equation}
self-generated trajectories to reach excess risk $\epsilon$. Crucially, the bound depends on the timing-head capacity and the encoder's \emph{relevance reliability} $\epsilon_V$, but \emph{not} on the encoder's overall capacity. This decoupling is what makes self-evolution work without external annotators.

\paragraph{Two falsifiable predictions.} Eq.~\eqref{eq:combined} predicts:
\begin{enumerate}[leftmargin=1.4em,topsep=2pt,itemsep=2pt]
  \item \textbf{Strong-encoder regime.} When the base model is a competent VideoLLM on the target domain, $\epsilon_V$ is small and the noise-inflation factor $1/(1-2\epsilon_V)^2$ stays close to $1$, so a small budget $n\approx 10^3$ already activates the timing policy. This matches the consistent gains we observe across Qwen2/2.5/3-VL, InternVL-3.5, and MiniCPM-V4.5 in Table~\ref{tab:main_exp}.
  \item \textbf{Weak-encoder regime.} As $\epsilon_V\to 1/2$ (e.g., medical or industrial video that is far from the encoder's pretraining distribution), the noise-inflation factor diverges and self-evolution must be paired with external supervision before becoming effective. This is consistent with the limitations we discuss in Section~\ref{sec:self_data_quality}, and is, to our knowledge, the first quantitative criterion for when self-evolved streaming data is or is not appropriate.
\end{enumerate}

\subsection{Analysis of Training Efficiency and Resolution Robustness}
\label{sec:resolution}
\begin{table*}[!t]
  \centering
  \setlength{\tabcolsep}{3.2pt}
  \renewcommand{\arraystretch}{1.18}
  \caption{\textbf{Impact of visual token density on RealStreamEval.} All EvoStreaming variants are trained at 128 tokens per frame; we evaluate at both 128 and 768 (full resolution). The efficient training setting generalizes well: switching to 768 tokens at inference consistently improves Real-Time Visual Perception while preserving timing behavior learned at low resolution.}
  \label{tab:token_ablation}
  \resizebox{\textwidth}{!}{%
  \begin{tabular}{@{}ll ccccccc cccc cccc c@{}}
  \toprule
  \multirow{2}{*}{\textbf{Model}} &
  \multirow{2}{*}{\textbf{Tok.}} &
  \multicolumn{7}{c}{\textbf{Real-Time Visual Perception}} &
  \multicolumn{4}{c}{\textbf{Backward Tracing}} &
  \multicolumn{4}{c}{\textbf{Forward Active Responding}} &
  \textbf{Overall} \\
  \cmidrule(lr){3-9} \cmidrule(lr){10-13} \cmidrule(lr){14-17} \cmidrule(lr){18-18}
  & & OCR & ACR & ATR & STU & FPD & OJR & Avg. &
  EPM & ASI & HLD & Avg. &
  REC & SSR & CRR & Avg. &
  Avg. \\
  \midrule
  \multicolumn{18}{@{}l}{\cellcolor{gray!6}\textit{\;\;Vanilla models}} \\
  \addlinespace[1pt]
  Qwen2-VL-8B
    & 128 & 60.7 & 56.4 & 56.0 & 31.2 & \textbf{50.5} & 56.5 & 51.9
          & 41.8 & \textbf{49.7} & 53.5 & 48.3
          & \textbf{16.3} & \textbf{21.2} & \textbf{1.6} & \textbf{13.1} & 41.3 \\
    & 768 & \textbf{75.2} & \textbf{58.3} & \textbf{61.2} & \textbf{33.7} & 50.0 & \textbf{59.8} & \textbf{56.4}
          & \textbf{48.2} & 44.3 & \textbf{57.0} & \textbf{49.8}
          & 16.0 & 21.1 & 0.8 & 12.6 & \textbf{43.8} \\
  \addlinespace[2pt]
  Qwen2.5-VL-8B
    & 128 & 60.7 & \textbf{56.4} & 56.0 & 31.2 & 50.5 & 56.5 & 51.9
          & \textbf{41.8} & 49.7 & \textbf{53.5} & \textbf{48.3}
          & \textbf{15.6} & \textbf{22.2} & 1.2 & \textbf{13.0} & 41.3 \\
    & 768 & \textbf{78.5} & 47.7 & \textbf{58.6} & \textbf{38.2} & \textbf{56.4} & \textbf{58.2} & \textbf{56.3}
          & 41.6 & \textbf{56.8} & 42.5 & 46.9
          & 14.9 & 15.0 & \textbf{8.0} & 12.6 & \textbf{43.0} \\
  \addlinespace[2pt]
  Qwen3-VL-8B
    & 128 & 49.0 & \textbf{45.0} & 56.0 & 41.6 & \textbf{42.1} & 53.3 & 47.8
          & \textbf{36.7} & \textbf{51.0} & 38.2 & \textbf{42.0}
          & \textbf{17.3} & 23.4 & 3.9 & 14.8 & 38.1 \\
    & 768 & \textbf{54.7} & 44.5 & \textbf{57.3} & \textbf{43.8} & 35.2 & \textbf{56.0} & \textbf{48.6}
          & 34.2 & 50.3 & 38.2 & 40.9
          & 17.1 & \textbf{24.6} & \textbf{5.6} & \textbf{15.8} & \textbf{38.5} \\
  \midrule
  \multicolumn{18}{@{}l}{\cellcolor{gray!6}\textit{\;\;\MethodName{} (ours)}} \\
  \addlinespace[1pt]
  \MethodName-Qwen2
    & 128 & 72.8 & 63.3 & 69.0 & 49.7 & 54.5 & 63.0 & 62.1
          & 48.5 & 54.7 & 40.9 & 48.0
          & \textbf{20.6} & 62.5 & \textbf{9.2} & \textbf{30.8} & 50.7 \\
  \rowcolor{SkyBlue!15}
    & 768 & \textbf{82.9} & \textbf{67.9} & \textbf{75.9} & \textbf{53.4} & \textbf{62.4} & \textbf{69.0} & \textbf{68.6}
          & \textbf{51.5} & \textbf{55.4} & \textbf{45.2} & \textbf{50.7}
          & 20.6 & \textbf{64.6} & 6.7 & 30.6 & \textbf{54.6} \\
  \addlinespace[2pt]
  \MethodName-Qwen2.5
    & 128 & 69.5 & \textbf{65.6} & \textbf{66.8} & 50.6 & \textbf{58.9} & \textbf{61.7} & \textbf{62.2}
          & \textbf{46.1} & 50.0 & \textbf{39.0} & \textbf{45.0}
          & \textbf{19.3} & \textbf{50.1} & 5.4 & \textbf{24.9} & \textbf{48.6} \\
  \rowcolor{SkyBlue!15}
    & 768 & \textbf{78.9} & 58.3 & 63.4 & \textbf{51.1} & 56.9 & 60.9 & 61.6
          & 45.3 & \textbf{53.0} & 35.2 & 44.5
          & 18.9 & 38.7 & \textbf{7.9} & 21.8 & 47.4 \\
  \addlinespace[2pt]
  \MethodName-Qwen3
    & 128 & 66.1 & 50.5 & 61.6 & 36.0 & 37.6 & 51.1 & 50.5
          & 35.9 & 28.0 & \textbf{36.8} & 33.6
          & 16.8 & 55.5 & 6.7 & 26.3 & 40.2 \\
  \rowcolor{SkyBlue!15}
    & 768 & \textbf{79.9} & 50.5 & \textbf{62.1} & \textbf{37.1} & \textbf{38.6} & \textbf{60.6} & \textbf{54.8}
          & \textbf{45.1} & \textbf{41.2} & 33.6 & \textbf{40.0}
          & \textbf{19.3} & \textbf{69.3} & \textbf{24.2} & \textbf{37.6} & \textbf{46.8} \\
  \bottomrule
  \end{tabular}%
  }
\end{table*}

In this section, we analyze the impact of visual token density on streaming performance. During the self-evolution training phase, we limit the visual input to $128$ tokens per frame to reduce computational overhead. During evaluation, we also test at the full $768$ token resolution (or the model's native maximum) to examine whether higher visual fidelity changes streaming behavior.

To validate the robustness of this strategy, we compare the performance of both vanilla models and our EvoStreaming-adapted models under two inference settings: $128$ tokens (matching training) and $768$ tokens (high resolution). The results are summarized in Table~\ref{tab:token_ablation}.

\paragraph{Performance Stability across Resolutions:}
As observed in Table~\ref{tab:token_ablation}, EvoStreaming models remain stable when evaluated with higher-resolution inputs ($768$ tokens) despite being trained with coarser visual representations ($128$ tokens). This suggests that the training signal, focused on ``when to speak'' and interaction timing, is not tightly tied to the specific token density used during training.

\paragraph{Benefit of Higher Resolution for Perception:}
For models with stronger visual encoders like \textbf{Qwen3-VL}, the \textbf{768-token} setting yields higher performance in the \textbf{Real-Time Visual Perception} tasks than the $128$-token setting. This indicates that while $128$ tokens can provide useful supervision for learning the streaming interaction policy, higher-resolution inference can still help the model recognize fine-grained visual details such as OCR and small objects.

\subsection{Quality and Limitations of Self-Generated Supervision}
\label{sec:self_data_quality}

A natural concern is whether self-generated trajectories provide reliable training labels, or whether the model simply reinforces its own preferences. Prior work shows that model-generated instruction data can be effective when carefully filtered or curated~\citep{wang2023selfinstruct,zhou2023lima}, but also warns that repeatedly training on generated data can amplify errors or distributional collapse when supervision is unconstrained~\citep{shumailov2024modelcollapse}. EvoStreaming mitigates this risk by decomposing data construction into constrained substeps rather than asking the model to produce unconstrained streaming conversations in one pass. Stages 1 and 2 determine the task type and generate type-consistent questions. Stage 3 then reduces temporal grounding to a binary relevance judgment over short video segments: whether the evidence needed for a question is visible in the current segment. Stage 4 converts these local relevance labels into causal trajectories with explicit \textsc{Silent} and \textsc{Respond} actions.

This design makes the supervision narrower than open-ended answer generation. The most important labels for streaming adaptation are timing labels, especially when to withhold output. These labels are lower-dimensional than full visual-language reasoning, and they rely on visual abilities that the base model has already learned during offline pre-training. The resulting data should therefore be viewed as a way to expose an interaction policy from the model's existing visual knowledge, not as a guarantee that all pseudo-labels are perfectly correct.

Self-preference bias is also limited by the evaluation setup. The reported gains are not measured by the same model that generated the data. They are evaluated under RealStreamEval using protocol-defined answer windows, repetition penalties, and external judge prompts. If self-evolution only amplified the model's original response style, it would not necessarily improve under a protocol that penalizes excessive or premature responses. The improvement in silent behavior and token efficiency suggests that the adaptation changes the interaction pattern rather than only reinforcing response wording.

\paragraph{Manual audit on $n=50$ trajectories.}
To empirically verify the data quality, we sampled $n=50$ trajectories uniformly from $\mathcal{D}$ (built on Qwen2-VL as the base model) and had two authors independently inspect each along the five quality dimensions in Table~\ref{tab:self_data_audit_protocol}; disagreements were resolved by discussion. Trajectory consistency reaches $100\%$, reflecting the deterministic Stage~3 mapping from the relevance matrix to actions, while the four stochastic dimensions stay above $80\%$. Prompt-conditioned stages, namely task type ($94\%$) and question answerability ($90\%$), are most reliable, and the hardest dimension is segment-level relevance judgement ($84\%$), reflecting the intrinsic difficulty of deciding whether a short clip contains the evidence needed for a self-generated question. Overall, the audit supports the assumption that self-generated supervision is of sufficient quality for streaming timing adaptation when the base model is a competent VideoLLM on the target domain.

\begin{table*}[!t]
\centering
\caption{\textbf{Manual audit of $n=50$ self-generated trajectories} (base model: Qwen2-VL). Two authors independently labelled each trajectory and resolved disagreements by discussion. The fourth column states the pipeline mechanism each criterion relies on or that bounds its failure mode; the third column reports the measured pass rate.}
\label{tab:self_data_audit_protocol}
\small
\begin{tabular}{p{0.18\textwidth}p{0.30\textwidth}p{0.16\textwidth}p{0.26\textwidth}}
\toprule
\textbf{Audit item} & \textbf{Acceptance criterion} & \textbf{Pass rate} & \textbf{Pipeline mechanism} \\
\midrule
Task type consistency & The generated question matches the assigned temporal category (IV, MD, TA, AM, or DED). & $94\%$ ($47/50$) & Stage~1/2 prompts explicitly condition on $c$ (Alg.~\ref{alg:evo} L4--5); type drift requires overriding the prompt. \\
Question answerability & The question can be answered from visual evidence in the video rather than requiring outside knowledge. & $90\%$ ($45/50$) & Stage~2 conditions on the video and category, not external knowledge; Stage~3 discards questions with no \textsc{Relevant} segments (Alg.~\ref{alg:evo} L6). \\
Temporal relevance quality & Segments labelled \textsc{Relevant} contain the evidence needed to answer the question, and irrelevant segments do not. & $84\%$ ($210/250$ pairs) & Stage~3 binary relevance judgement on short segments; quality is bounded by the base encoder's recognition ability on the target video distribution. \\
Trajectory consistency & The derived \textsc{Silent}/respond actions follow the relevance matrix under the causal streaming constraint. & $100\%$ ($50/50$) & Stage~4 is a deterministic function of $\mathbf{A}$ (Alg.~\ref{alg:evo} L7--12); satisfied by construction. \\
Sample validity & Samples are not malformed, unanswerable, temporally inconsistent, or duplicate near-identical trajectories. & $4\%$ fail ($2/50$) & Stage~5 filters before fine-tuning (Alg.~\ref{alg:evo} L13). \\
\bottomrule
\end{tabular}
\end{table*}

The main limitation is domain dependence. If the base model's visual perception is unreliable in a target domain, such as specialized medical or industrial video, the relevance labels produced by self-annotation may become noisy. In such settings, EvoStreaming should be combined with domain-specific visual adaptation or a small amount of expert-labeled temporal supervision. We therefore view self-evolution as most appropriate when the base VideoLLM already has reasonable visual understanding on the target video distribution.
\subsection{Why Stronger Offline Models Do Not Always Score Higher}
\label{sec:verbosity_analysis}

RealStreamEval can produce model rankings that differ from standard offline video benchmarks. This does not necessarily imply that newer or stronger backbones have weaker visual understanding. One plausible explanation is that streaming evaluation couples answer quality with response timing and response frequency. A model that tends to produce detailed outputs at many timesteps may be strong in offline QA but less suitable for a streaming assistant that should often remain silent.

To analyze this behavior, we report a \textbf{Per-Token Score} ($\eta$), defined as the ratio of the model's overall accuracy to the average number of tokens generated per turn during active monitoring:
\begin{equation}
\eta = \frac{\text{Overall Score (\%)}}{\text{Avg. Tokens}}
\end{equation}
This metric serves as a proxy for \textit{information density}: it quantifies how much score the model obtains per generated token. It is not intended to replace task accuracy, but it helps reveal whether a model's streaming score is affected by excessive generation.

\begin{table*}[!t]
  \centering
  \caption{\textbf{Token-level analysis of streaming behavior.} We report the Per-Token Score ($\eta$), a proxy for the information density of each model under RealStreamEval. Lower $\eta$ often reflects verbose or redundant outputs during timesteps where a streaming assistant should remain silent.}
  \label{tab:per_token_score}
  \resizebox{0.8\linewidth}{!}{%
  \scriptsize
  \begin{tabular}{lcccc}
  \toprule
  \textbf{Model} & \textbf{Acc. (\%)} & \textbf{Avg. Tokens} & \textbf{Score ($\eta$)} & \textbf{Gain} \\
  \midrule
  \textit{Vanilla Baselines} & & & & \\
  Qwen3-VL-8B & 38.5 & 283.47 & 0.14 & - \\
  Qwen2.5-VL-8B & 43.0 & 63.78 & 0.67 & - \\
  Qwen2-VL-8B & 43.8 & 51.21 & 0.86 & - \\
  \midrule
  \textit{EvoStreaming (Ours)} & & & & \\
  \MethodName-8B-Qwen3 & \textbf{46.8} & 10.25 & \textbf{4.56} & \textbf{33.6$\times$} \\
  \MethodName-8B-Qwen2.5 & \textbf{47.4} & \textbf{8.18} & \textbf{5.79} & \textbf{8.6$\times$} \\
  \MethodName-8B-Qwen2 & \textbf{54.6} & 8.45 & \textbf{6.46} & \textbf{7.5$\times$} \\
  \bottomrule
  \end{tabular}%
  }
\end{table*}

\paragraph{Verbose Outputs Can Hurt Streaming Scores.}
As detailed in Table~\ref{tab:per_token_score}, vanilla offline models exhibit low token efficiency under RealStreamEval. For instance, Qwen3-VL achieves a Per-Token Score of 0.14 while generating 283.47 tokens per turn on average. This pattern is consistent with a verbosity mismatch: a model may provide detailed descriptions or repeated observations when the protocol expects silence, which increases the answer rate and activates the verbosity penalty.

\paragraph{Effect of EvoStreaming.}
After EvoStreaming adaptation, all three Qwen-family backbones generate far fewer tokens per turn, around 8 to 10 on average, while improving their RealStreamEval scores. Qwen3-VL is the clearest example: its average token count drops from 283.47 to 10.25, and its Per-Token Score increases from 0.14 to 4.56. These results suggest that one important effect of EvoStreaming is not only improving answer content, but also teaching the model to withhold output when the stream does not require a response.

\section{Data prompt}
\section{Prompts for Self-Evolved Data Construction}
\label{app_sec:data_generation_prompts}

This section reports the prompt design used to construct the self-evolved trajectories in Algorithm~\ref{alg:evo}. The implementation factors the prompts into three stages that match the data pipeline in Fig.~\ref{fig:data_pipeline}: task-aware self-question generation, segment-level relevance annotation, and causal roll-out into \textsc{Silent}/respond trajectories. The prompts are organized according to the five task categories introduced in Section~\ref{sec:evostreaming}, as summarized in Table~\ref{tab:data_prompt_categories}.

\begin{table*}[!ht]
\centering
\setlength{\tabcolsep}{6pt}
\renewcommand{\arraystretch}{1.12}
\caption{\textbf{Task categories used by the self-evolved data-generation prompts.} Each category induces a different type of streaming supervision, ranging from immediate visual grounding to patient monitoring and proactive response.}
\label{tab:data_prompt_categories}
\small
\begin{tabular}{@{}>{\raggedright\arraybackslash}p{0.34\textwidth}>{\raggedright\arraybackslash}p{0.28\textwidth}>{\raggedright\arraybackslash}p{0.3\textwidth}@{}}
\toprule
\textbf{Task category} & \textbf{Prompt objective} & \textbf{Streaming behavior elicited} \\
\midrule
Immediate Visual (IV) & Generate questions about moment-specific visible details & Ground responses in current visual evidence \\
Memory-Dependent (MD) & Generate recall questions about earlier events & Retrieve relevant past observations when queried later \\
Temporal Aggregation (TA) & Generate counting questions over repeated actions & Accumulate evidence across repeated events \\
Anticipatory Monitoring (AM) & Generate questions about hidden targets or future reveals & Wait until visual evidence becomes verifiable \\
Dynamic Event Description (DED) & Generate questions about sequential steps or procedures & Report meaningful progress in evolving activities \\
\bottomrule
\end{tabular}
\end{table*}

\newenvironment{dataprompttable}[3]{
  \begin{table*}[!ht]\centering
  \caption{#2}
  \label{#3}
  \begin{minipage}{1.0\textwidth}
  \centering
  \begin{tcolorbox}
  \footnotesize
  \renewcommand{\arraystretch}{1.1}
  \begin{tabular}{p{\dimexpr\linewidth-2\tabcolsep\relax}}
  \VarSty{{\bf #1}}\\[1mm]
}{
  \end{tabular}
  \end{tcolorbox}
  \end{minipage}
  \end{table*}
}

\subsection{Stage 1: Task-Aware Self-Question Generation}
\label{app_sec:data_prompt_stage1}

Stage~1 asks the base model to watch the full video and generate task-consistent questions. The shared structure is shown below; the task-specific clause is selected according to the category assigned by $\mathcal{M}_{\mathrm{cls}}$.

\begin{dataprompttable}{Prompt B.1: Full-video Self-Question Generation}{Prompt template used for task-aware full-video question generation in self-evolved data construction.}{tab:data_prompt_stage1}
\textbf{Role.} You are an expert video analyst specializing in \{task-specific video skill\}. Watch the video carefully and completely from start to finish. Do not rush; pay attention to details across the entire timeline.\\[1mm]
\textbf{Task.} Generate high-quality questions based only on what is actually visible in the video. The questions should be natural, specific to the video content, and answerable from visual evidence.\\[1mm]
\textbf{Task-specific instruction.}
\begin{itemize}[leftmargin=1.2em,topsep=1pt,itemsep=1pt]
    \item \textbf{Immediate Visual (IV):} generate diverse questions about specific visual details at the beginning, middle, and end of the video, including objects, people, actions, colors, text, numbers, counts, positions, and states.
    \item \textbf{Memory-Dependent (MD):} generate recall questions that test memory of earlier events from the perspective of later timestamps, using cues such as ``Earlier'', ``Remember'', ``At the beginning'', ``Before'', or ``Previously''.
    \item \textbf{Temporal Aggregation (TA):} generate counting questions about distinct, countable repetitive actions with clear boundaries, such as jumps, claps, steps, taps, cuts, or object placements.
    \item \textbf{Anticipatory Monitoring (AM):} generate questions about hidden targets or future reveals, where an action, gaze, reach, opening, reaction, or camera motion creates a setup and later frames reveal the answer.
    \item \textbf{Dynamic Event Description (DED):} generate questions about sequential steps, procedures, or multi-step workflows, emphasizing what happens and in what order.
\end{itemize}
\textbf{Output format.} First describe the observed video content in one or two sentences. Then list questions as \texttt{Q1: ...}, \texttt{Q2: ...}, \texttt{Q3: ...}.
\end{dataprompttable}

\subsection{Stage 2: Segment-Level Relevance Annotation}
\label{app_sec:data_prompt_stage2}

Stage~2 applies the base model to short video segments and asks whether each segment contains evidence for each generated question. In our implementation, each segment prompt is specialized to the task family so that relevance is judged with the correct temporal semantics.

\begin{dataprompttable}{Prompt B.2: Segment-Level Visual Evidence Annotation}{Prompt template used to annotate whether each short segment contains visual evidence for each generated question.}{tab:data_prompt_stage2}
\textbf{Input.} A short video segment, the current segment time range, and the generated questions \{\texttt{questions\_text}\}.\\[1mm]
\textbf{Shared instruction.} Analyze only the visible evidence in this segment. For each question, determine whether the segment contains information relevant to answering or updating that question. Do not infer from off-screen or future content.\\[1mm]
\textbf{Task-specific annotation target.}
\begin{itemize}[leftmargin=1.2em,topsep=1pt,itemsep=1pt]
    \item \textbf{Immediate Visual (IV):} output whether the requested attribute is clearly visible; if yes, provide the exact evidence and best-view time window.
    \item \textbf{Memory-Dependent (MD):} output whether this segment depicts the earlier origin event requested by the recall question; if yes, log the event and timestamp.
    \item \textbf{Temporal Aggregation (TA):} distinguish complete repetitions from partial repetitions; report completed cycle counts, timing, and partial-start or partial-end notes.
    \item \textbf{Anticipatory Monitoring (AM):} classify the segment as \textsc{Setup}, \textsc{Reveal}, \textsc{Post-Reveal}, or \textsc{N/A}; the reveal is the first moment when the answer becomes visually verifiable.
    \item \textbf{Dynamic Event Description (DED):} determine whether the segment shows execution of the queried procedural step; if yes, report the micro-action, phase, and timing.
\end{itemize}
\textbf{Output format.} For each question, return a structured line such as \texttt{- Question 1: [Yes/No or Class] - [Evidence, timing, count, phase, or N/A]}. Segments for which all questions are \textsc{No}/\textsc{N/A} are filtered out before roll-out.
\end{dataprompttable}

\subsection{Stage 3: Question Selection and Causal Roll-Out}
\label{app_sec:data_prompt_stage3}

Stage~3 converts segment-level annotations into training trajectories. The prompt first selects a trackable question with sufficient evidence, then repeatedly decides whether the current accumulated window warrants a response or should remain \textsc{Silent}. This stage supplies the timing labels used for fine-tuning.

\begin{dataprompttable}{Prompt B.3: Selecting a Tracked Question}{Prompt template used to select a question with sufficient segment-level evidence for causal roll-out.}{tab:data_prompt_stage3_select}
\textbf{Role.} You are selecting the best question to track for this video.\\[1mm]
\textbf{Input.} Task type \{task\_type\}, numbered available questions, and sample captions from the first several annotated segments.\\[1mm]
\textbf{Instruction.}
\begin{enumerate}[leftmargin=1.4em,topsep=1pt,itemsep=1pt]
    \item Analyze the sample captions and identify which question has the most relevant \textsc{Yes} evidence across segments.
    \item Prefer questions with concrete, observable information rather than ambiguous or \textsc{N/A} evidence.
    \item For Temporal Aggregation, prefer repeated actions with count information; for Dynamic Event Description, prefer step-by-step processes; for Anticipatory Monitoring, prefer changing states or reveal chains.
\end{enumerate}
\textbf{Output format.} Return JSON with \texttt{selected\_question\_idx}, an initial \texttt{task\_prompt} for tracking that question, and a brief \texttt{reasoning} field.
\end{dataprompttable}

\begin{dataprompttable}{Prompt B.4: Streaming Response Decision for a Tracked Question}{Prompt template used to convert accumulated segment evidence into a \textsc{Silent} or response action.}{tab:data_prompt_stage3_decide}
\textbf{Role.} You are analyzing video segments to decide whether to generate a response.\\[1mm]
\textbf{Input.} Current task type \{task\_type\}, the tracked question number and text, current timestamp, accumulated captions in the current time range, last response, and full conversation history.\\[1mm]
\textbf{Instruction.}
\begin{enumerate}[leftmargin=1.4em,topsep=1pt,itemsep=1pt]
    \item Focus only on annotation lines corresponding to the tracked question.
    \item Check the last response to avoid repeating information already reported.
    \item Decide whether there is enough new evidence to produce an update; otherwise return \textsc{Silent}.
    \item Balance responsiveness and sparsity: avoid spamming, but do not stay silent if the user would miss a key count, step transition, or reveal by waiting.
\end{enumerate}
\textbf{Task-specific response rule.}
\begin{itemize}[leftmargin=1.2em,topsep=1pt,itemsep=1pt]
    \item \textbf{Temporal Aggregation (TA):} respond when one or more new repetitions are completed, maintaining the running count from the last response.
    \item \textbf{Dynamic Event Description (DED):} respond on step transitions or meaningful progress within a step; avoid updates when the same step continues without new information.
    \item \textbf{Anticipatory Monitoring (AM):} respond when new visual evidence significantly changes the current understanding, progressing from ambiguity to hypothesis to confirmation.
    \item \textbf{Immediate Visual (IV) and Memory-Dependent (MD):} ask or answer only when clear visual evidence exists in the current window; Memory-Dependent questions must refer to earlier events at least roughly one minute in the past.
\end{itemize}
\textbf{Output format.} Return JSON with \texttt{should\_respond}, a short \texttt{reason}, and \texttt{response} only when \texttt{should\_respond=true}. A false decision is converted to the \textsc{Silent} action in the training trajectory.
\end{dataprompttable}

\subsection{Why CRR Remains a Bottleneck}
\label{sec:crr_bottleneck}

Within Forward Active Responding (FAR), the Clue-based Reasoning Response (CRR) sub-task asks the model to wait until visual evidence emerges and then produce a substantive causal answer about \emph{why} that evidence implies a particular event. CRR is the most demanding of the three FAR sub-tasks: the FAR advantage of newer Qwen versions over older ones breaks down to a single dominant factor, namely that Qwen3-VL's CRR lead over Qwen2-VL ($24.2$ versus $6.7$) drives most of its FAR advantage. Yet across all systems, including ours, CRR remains the lowest-scoring sub-task: our best configuration reaches $29.1\%$ (InternVL3.5) while human performance is $93.5\%$. Two factors clarify the gap. CRR is the only Forward Active Responding task scored with a \emph{strict} intention judge (Table~\ref{tab:prompt_crr_intention}): the model must not only respond at the right moment but also produce a substantive clue-based causal reply, whereas SSR and REC use a lenient judge that admits descriptive or recognitive intents. EvoStreaming primarily supervises \emph{when} to respond, through binary relevance labels and the \textsc{Silent}/respond action space, and does not directly shape the content of triggered responses. We therefore read CRR not as a timing failure but as a content-quality ceiling, and as a concrete direction for future work: combining timing supervision from self-evolved trajectories with rationale- or process-based content rewards, without falling back on large-scale streaming annotation.

\section{Evaluation Prompts for RealStreamEval}
\label{appendix:prompts}

This section details the specific prompts used by the LLM-as-a-judge to evaluate model performance and penalize undesired behaviors within the RealStreamEval framework. The prompts are grouped by their corresponding evaluation components: answer correctness in Table~\ref{tab:prompt_perception_tracing}, repetition detection in Table~\ref{tab:prompt_repetition}, response intention detection in Tables~\ref{tab:prompt_crr_intention} and~\ref{tab:prompt_ssr_rec_intention}, and Forward Active Responding consistency in Tables~\ref{tab:prompt_far_crr}, \ref{tab:prompt_far_ssr}, and~\ref{tab:prompt_far_rec}.

\newenvironment{prompttable}[3]{
  \begin{table*}[!ht]\centering
  \caption{#2}
  \label{#3}
  \begin{minipage}{1.0\textwidth}
  \centering
  \begin{tcolorbox}
  \footnotesize
  \renewcommand{\arraystretch}{1.1}
  \begin{tabular}{p{\textwidth}}
  \VarSty{{\bf #1}}\\[1mm]
}{
  \end{tabular}
  \end{tcolorbox}
  \end{minipage}
  \end{table*}
}

\subsection{LLM Accuracy Judge for Perception and Tracing Tasks}
\label{app_sec:prompt_perception_tracing}
Used for \textit{Real-Time Visual Perception} and \textit{Backward Tracing} tasks. The exact judge prompt for this component is shown in Table~\ref{tab:prompt_perception_tracing}.

\begin{prompttable}{Prompt C.1: Accuracy Judge for Perception and Tracing}{Prompt template used to judge answer correctness for Real-Time Visual Perception and Backward Tracing tasks.}{tab:prompt_perception_tracing}
You are an expert evaluator for video understanding tasks. \\
Task: \{task\_name\} (\{task\}) \\
Question: \{question\} \\
Options: \{options\} \\
Ground Truth Answer: \{ground\_truth\} \\
Model's Answer: \{model\_answer\} \\
Your task:
\begin{enumerate}
    \item Carefully compare the model's answer with the ground truth answer.
    \item Determine if the model's answer is correct.
    \item For multiple choice questions, check if the model selected the correct option (either by letter or by content).
    \item For open-ended questions, check if the model's answer captures the same meaning as the ground truth.
\end{enumerate}
Respond with a JSON object in exactly this format:
\{ "correct": true or false, "reasoning": "Brief explanation of your judgment" \} \\
Only output the JSON object, nothing else.
\end{prompttable}

\subsection{Repetition Detection Judge for Penalty}
\label{app_sec:prompt_repetition}
This prompt is used to implement the repetition penalty within the $\mathcal{M}(R_{ans})$ mechanism. The corresponding prompt template is provided in Table~\ref{tab:prompt_repetition}.

\begin{prompttable}{Prompt C.2: Repetition Detection Judge}{Prompt template used to detect repetitive responses for the RealStreamEval repetition penalty.}{tab:prompt_repetition}
You are checking for repetitive behavior in an AI agent's responses. \\
Context (User/Model dialogue turns, truncated): \{context\_text\} \\
Ground Truth Answer: \{ground\_truth\} \\
Task:
\begin{enumerate}
    \item Analyze the Context provided above.
    \item Determine if the agent repeats the answer to the question multiple times unnecessarily within this context.
    \item If the answer appears more than once (e.g., 'It is red. I see red. It is red'), mark it as repeated.
    \item If the agent answers once and then stays silent or moves to the next topic, it is NOT repeated.
\end{enumerate}
Respond with a JSON object in exactly this format:
\{ "is\_repeated": true or false, "reasoning": "Why you think it is repeated or not" \} \\
Only output the JSON object.
\end{prompttable}

\subsection{Intention Judgment Prompts}
\label{app_sec:prompt_intention}
Used to detect whether a model's output constitutes a valid response intention. The strict CRR intention judge is shown in Table~\ref{tab:prompt_crr_intention}, while the lenient SSR/REC intention judge is shown in Table~\ref{tab:prompt_ssr_rec_intention}.

\subsubsection{CRR Task (Strict Judgment)}
\label{app_sec:prompt_crr_intention}
Table~\ref{tab:prompt_crr_intention} gives the strict intention prompt used for CRR outputs.

\begin{prompttable}{Prompt C.3: Strict Intention Judge for CRR}{Prompt template used to determine whether a CRR output contains a substantive response intention.}{tab:prompt_crr_intention}
Question: \{question\} \\
Model's Output: \{content\} \\
Does this output provide a \textbf{substantive and direct} ANSWER or RESPONSE to the question? \\
Consider as a valid intention (answer "yes") ONLY if:
\begin{itemize}
    \item The output makes a direct attempt to answer the question posed.
    \item It provides specific details, reasoning or observations that are RELEVANT to the question.
\end{itemize}
DO NOT consider as having valid intention (answer "no") if the output is vague, rephrases the question, or is a simple refusal. \\
Respond with ONLY "yes" or "no".
\end{prompttable}

\subsubsection{SSR/REC Task (Lenient Judgment)}
\label{app_sec:prompt_ssr_rec_intention}
Table~\ref{tab:prompt_ssr_rec_intention} gives the lenient intention prompt used for SSR and REC outputs.

\begin{prompttable}{Prompt C.4: Lenient Intention Judge for SSR/REC}{Prompt template used to detect descriptive or recognition intent for SSR and REC tasks.}{tab:prompt_ssr_rec_intention}
Model's Output: \{content\} \\
Does this output show intention to DESCRIBE or RECOGNIZE something? \\
Consider as having intention (be lenient): Any description of actions, objects, or scenes; observations about what is visible. \\
Only consider as NOT having intention if completely empty or a pure refusal. \\
Respond with ONLY "yes" or "no".
\end{prompttable}
\subsection{LLM Accuracy Judge for Forward Active Responding}
\label{app_sec:prompt_far}
These prompts are used to calculate the Task Consistency Score for the \textit{Forward Active Responding} tasks, including CRR, SSR, and REC. The CRR, SSR, and REC judge prompts are listed in Tables~\ref{tab:prompt_far_crr}, \ref{tab:prompt_far_ssr}, and~\ref{tab:prompt_far_rec}, respectively.

\subsubsection{LLM Accuracy Judge Prompt for CRR Task}
\label{app_sec:prompt_far_crr}
Table~\ref{tab:prompt_far_crr} shows the CRR consistency judge used for Forward Active Responding.

\begin{prompttable}{Prompt C.5: Forward Active Responding Judge for CRR}{Prompt template used to score CRR responses in Forward Active Responding.}{tab:prompt_far_crr}
Question: \{question\} \\
Expected Answer: \{answer\} \\
Model's Response: \{prediction\} \\
Your task:
\begin{enumerate}
    \item Evaluate how well the model's response matches the expected answer.
    \item Respond with ONLY a score between 0.0 and 0.5, where:
    \begin{itemize}
        \item 0.5 = Good match with minor differences
        \item 0.3 = Related but somewhat differences
        \item 0.0 = Completely wrong or irrelevant
    \end{itemize}
\end{enumerate}
Only output the numerical score.
\end{prompttable}

\subsubsection{LLM Accuracy Judge Prompt for SSR Task}
\label{app_sec:prompt_far_ssr}
Table~\ref{tab:prompt_far_ssr} shows the SSR stage consistency judge used for Forward Active Responding.

\begin{prompttable}{Prompt C.6: Forward Active Responding Judge for SSR}{Prompt template used to score SSR stage consistency in Forward Active Responding.}{tab:prompt_far_ssr}
Expected Step/Stage: \{reference\} \\
Model's Response: \{prediction\} \\
\textbf{IMPORTANT:} For SSR, the model must correctly identify the STAGE/PHASE of the activity. Pay special attention to whether the model identifies the correct stage name or phase. \\
Your task:
\begin{enumerate}
    \item Respond with ONLY a score between 0.0 and 0.5, where:
    \begin{itemize}
        \item 0.5 = Perfect stage match, fully consistent
        \item 0.2 = Wrong stage but somewhat related activity
        \item 0.0 = Completely wrong stage or irrelevant
    \end{itemize}
\end{enumerate}
Only output the numerical score.
\end{prompttable}

\subsubsection{LLM Accuracy Judge Prompt for REC Task}
\label{app_sec:prompt_far_rec}
Table~\ref{tab:prompt_far_rec} shows the REC count consistency judge used for Forward Active Responding.

\begin{prompttable}{Prompt C.7: Forward Active Responding Judge for REC}{Prompt template used to score REC count consistency in Forward Active Responding.}{tab:prompt_far_rec}
Activity being counted: \{activity\} \\
Expected count at this point: \{expected\_count\} \\
Model's Response: \{prediction\} \\
Your task:
\begin{enumerate}
    \item Determine if the model correctly reports that the activity has occurred \{expected\_count\} time(s).
    \item Respond with ONLY a score between 0.0 and 0.5, where:
    \begin{itemize}
        \item 0.5 = Count is explicitly correct
        \item 0.3 = Count is approximately correct (off by 1)
        \item 0.0 = No count mentioned or completely wrong
    \end{itemize}
\end{enumerate}
Only output the numerical score.
\end{prompttable}

\section{Reliability and Scope of RealStreamEval}
\label{app_sec:reliability_scope}

RealStreamEval uses an LLM judge to map open-form model responses to predefined scoring criteria. We use Qwen3-VL-235B-A22B-Instruct~\citep{Qwen3-VL} served via vLLM~\citep{kwon2023efficient} as the judge across all experiments. This choice introduces a dependency on the judge model, but the judge is deliberately used in a constrained role. It is not asked to provide an open-ended preference judgment over response style. Instead, each prompt specifies the task type, ground-truth answer or expected event state, response timing rule, and output format. The judge therefore acts as a semantic executor of a rubric: it determines whether a candidate response matches the expected answer within the protocol-defined temporal context, and whether repeated responses should be penalized. The exact prompt templates for these decisions are provided in Appendix~\ref{appendix:prompts}.

The protocol also separates evaluation standards from judge implementation. The standards are defined by RealStreamEval itself: frame order, available history, response window, answer correctness, intention detection, and repetition penalty. The LLM judge is only used where exact string matching would be too brittle for natural-language answers. This design reduces the role of subjective preference, especially compared with free-form helpfulness or style evaluation. 

This rubric-first design is directly aligned with the LLM-as-a-judge paradigm that has been empirically validated in prior work. Initial research demonstrated that LLM-based evaluations can be consistent with expert human judgments when given the same instructions used for human annotators \citep{chiang2023can}. Subsequent studies further revealed that structured prompts with explicit evaluation criteria and chain-of-thought reasoning substantially improve alignment with human evaluation on NLG tasks \citep{liu2023g}. Similarly, investigations into LLM-as-a-judge frameworks using structured, rubric-based protocols report high agreement with human judgments~\citep{zheng2023judging}. Our prompts in Appendix~\ref{appendix:prompts} implement exactly the ingredients these works identify as decisive for judge reliability: (i) task-specific, fine-grained criteria rather than generic helpfulness scoring; (ii) explicit ground-truth answers, temporal windows, and intention rules that anchor the decision; (iii) chain-of-thought reasoning with structured output fields; and (iv) a separation between the evaluation protocol (defined by us) and the scoring operator (the LLM), so that the LLM serves as a semantic executor of a fixed rubric rather than a free-form judge. The empirical robustness observed in Appendix~\ref{sec:judge_robustness}, where the Overall ranking across five different LLM judges yields a mean pairwise Spearman of $\rho = 0.925$, is consistent with this literature: once the protocol is sufficiently specified, the identity of the LLM judge becomes a secondary factor.

Finally, RealStreamEval should be interpreted as a frame-indexed algorithmic evaluation rather than a complete deployment simulator. In our experiments, videos are sampled at 0.5 fps, so consecutive evaluation steps correspond to a two-second logical interval. This setting provides a clear and reproducible timeline for testing whether models can decide when to remain silent and when to respond. We do not directly score wall-clock latency, asynchronous queueing, or multi-user scheduling. These serving-system issues are important for deployment, but they are complementary to the model-level timing policy studied in this work.

\subsubsection{Judge Robustness}
\label{sec:judge_robustness}

We re-run the full RealStreamEval protocol on five representative models with four alternative proprietary judges in addition to the original Qwen3-VL-235B-A22B-Instruct~\citep{Qwen3-VL} judge: GPT-4o, GPT-5 mini, Claude-Haiku-4.5, and Gemini-2.5-Flash. All five judges are executed through the same rubric-constrained prompts described in Appendix~\ref{appendix:prompts}. 

Table~\ref{tab:crossjudge_overall_ranking} reports the Overall-score rank of five models under each of the five judges. The ranking is highly stable across judges: \MethodName-Qwen2 is ranked first by all five judges, MiniCPM-V4.5-8B is ranked last by all five judges, and Qwen2-VL-8B is ranked second under every judge. Quantitatively, the pairwise Spearman rank correlation between the paper judge (Qwen3-VL-235B-A22B) and each of the four alternative judges is $\rho\!\ge\!0.900$ (GPT-5 mini: $\rho\!=\!1.000$), with a mean of $\rho\!=\!0.925$; the minimum pairwise Spearman across \emph{all} judge pairs is also $\rho\!=\!0.900$. This indicates that the relative ordering of models produced by RealStreamEval is robust to the specific judge, and that the superiority of \MethodName\ over the corresponding offline baselines does not depend on the identity of the judge model. Together with the rubric-constrained protocol discussed in Appendix~\ref{app_sec:reliability_scope}, this reinforces the interpretation of the judge as a semantic executor of a fixed evaluation standard rather than the source of that standard.


\begin{table*}[!t]
  \centering
  \caption{\textbf{Judge-robustness test under RealStreamEval.} Per-judge Overall rank of five models under five different LLM judges.}
  \label{tab:crossjudge_overall_ranking}
  \setlength{\tabcolsep}{6pt} 
  \footnotesize 
  \begin{tabular}{lcccccc}
    \toprule
    Model & \makecell{Qwen3-VL-\\235B-A22B} & \makecell{Claude-\\Haiku-4.5} & \makecell{Gemini-2.5-\\Flash} & GPT-4o & \makecell{GPT-5\\mini} & \makecell{Avg.\\Rank} \\
    \midrule
    \makecell[l]{\MethodName-Qwen2}      & 1 & 1 & 1 & 1 & 1 & 1.00 \\
    \makecell[l]{Qwen2-VL-8B}            & 2 & 2 & 2 & 2 & 2 & 2.00 \\
    \makecell[l]{Qwen3-VL-8B}            & 4 & 3 & 3 & 3 & 4 & 3.40 \\
    \makecell[l]{\MethodName-MiniCPM-V}  & 3 & 4 & 4 & 4 & 3 & 3.60 \\
    \makecell[l]{MiniCPM-V4.5-8B}        & 5 & 5 & 5 & 5 & 5 & 5.00 \\
    \bottomrule
  \end{tabular}
\end{table*}

\section{Limitations and Future Works}
\label{sec:limitations}

RealStreamEval measures logical response timing, not wall-clock deployment latency, which depends on serving infrastructure and is orthogonal to the timing policy studied here. More substantially, EvoStreaming inherits the context window and visual reliability of its base model. To make this dependence quantitative, our analysis in Appendix~\ref{sec:theory} shows that the cost of self-evolved adaptation scales with $1/(1-2\epsilon_V)^2$ (Eq.~\eqref{eq:combined}), where $\epsilon_V$ is the encoder's segment-level relevance error. The strong-encoder regime ($\epsilon_V \ll 1/2$) covers the five mainstream VideoLLMs we evaluate. Outside this regime, for example on very long streams that exceed the context window or on specialized domains such as medical or industrial video that lie far from the encoder's pre-training distribution, the noise-inflation factor diverges, and self-evolution must be paired with external supervision (memory retrieval for long streams, or a small amount of expert temporal annotation for out-of-distribution domains) to recover effectiveness. A separate scope limitation is modality coverage: our current formulation is visual only, so audio cues such as speech activity and sound events fall outside the relevance matrix as defined.

\noindent\textbf{Future work.}
A natural next step is \emph{video reinforcement learning}, treating each streaming session as a Markov decision process in which the model observes a video segment, chooses \textsc{Silent} or \textsc{Respond}, and receives a dense reward that fuses answer correctness, response timing (softly penalizing both premature and delayed utterances), and a verbosity penalty. Because our supervised self-evolution already produces a strong initialization, we expect that a modest amount of online RL, possibly on the order of hundreds of streaming sessions, could push the timing policy beyond the static quality ceiling of self-generated labels and adapt it to domain-specific latency budgets in deployment.
A second direction is to close the CRR content-quality gap identified in Section~\ref{sec:analysis}. The current \textsc{Silent}/respond action space supervises \emph{when} to speak but not \emph{what} to say; integrating process-level rewards, such as answer-rationale consistency judged by a stronger model or fine-grained sub-question decomposition scores, could lift the weakest sub-task without requiring large-scale human annotation.
A third direction is to broaden the input modality and stream length: extending the relevance matrix with additional rows for speech activity and sound events would lift the visual-only restriction noted above, and pairing EvoStreaming with lightweight retrieval-augmented memory banks would let the framework handle streams that exceed the base model's context window.

\section{Case Studies}
\label{sec:case_studies}
\begin{figure*}[!t]
    \centering
    \includegraphics[width=\linewidth]{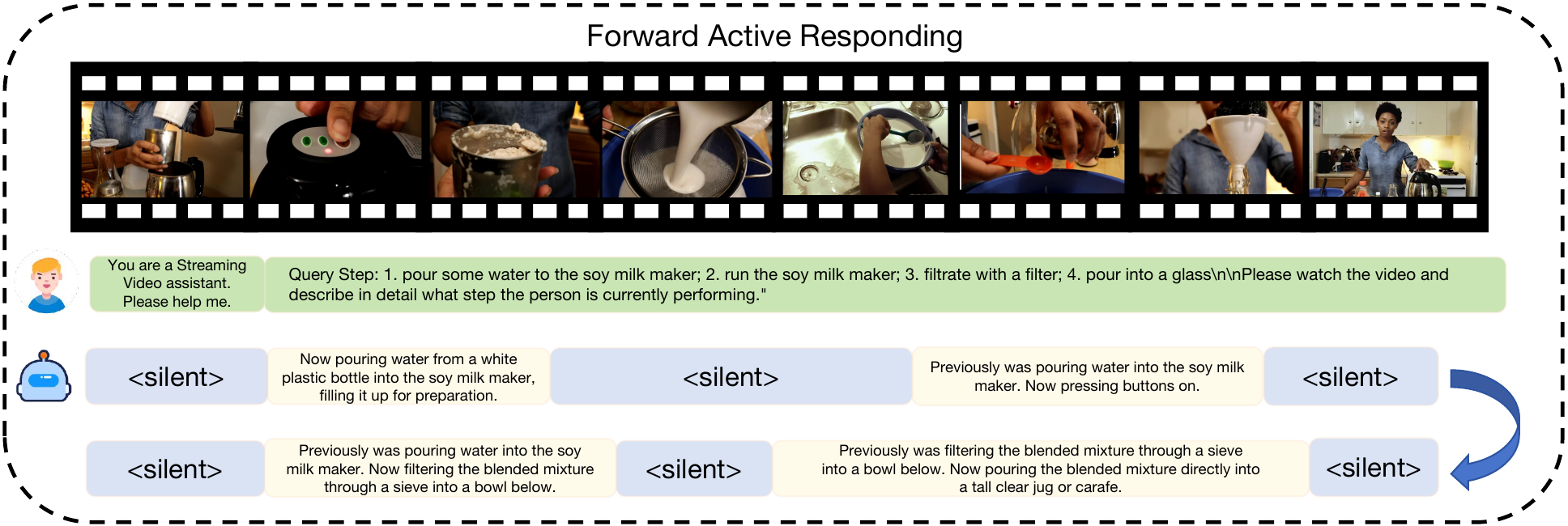}
    \caption{A case of Forward Active Responding. The model remains silent during most frames and generates detailed descriptions only when the person transitions to a new step in the soy milk preparation process.}
    \label{fig:case1}
\end{figure*}
\begin{figure*}[!t]
    \centering
    \includegraphics[width=\linewidth]{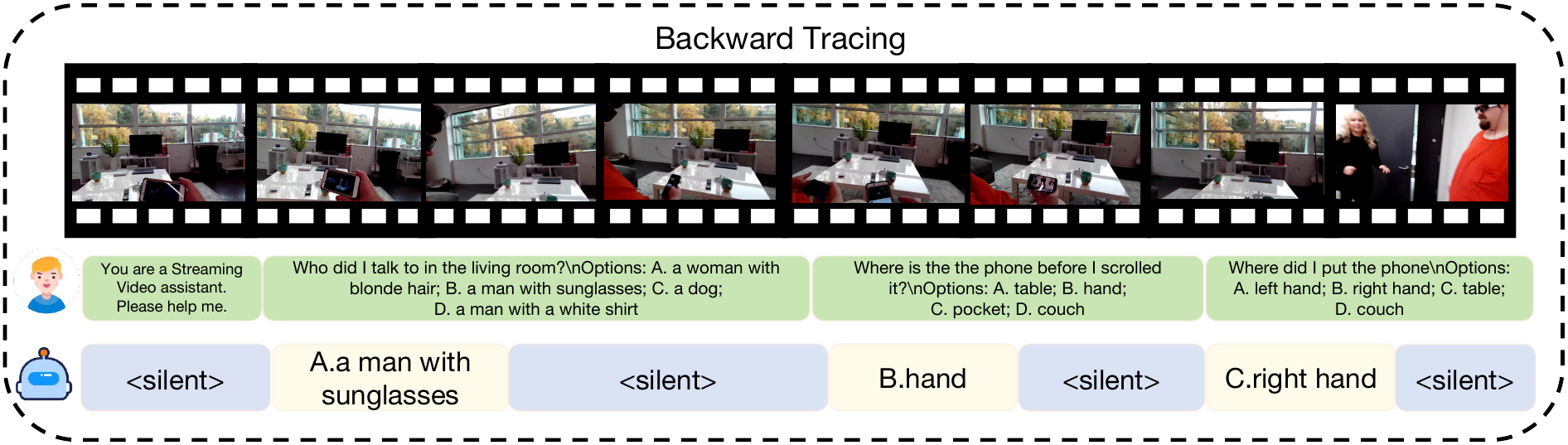}
    \caption{A case of Backward Tracing. The model maintains silence while observing, then accurately retrieves information from memory when queried about past events.}
    \label{fig:case2}
\end{figure*}
\begin{figure*}[!t]
    \centering
    \includegraphics[width=\linewidth]{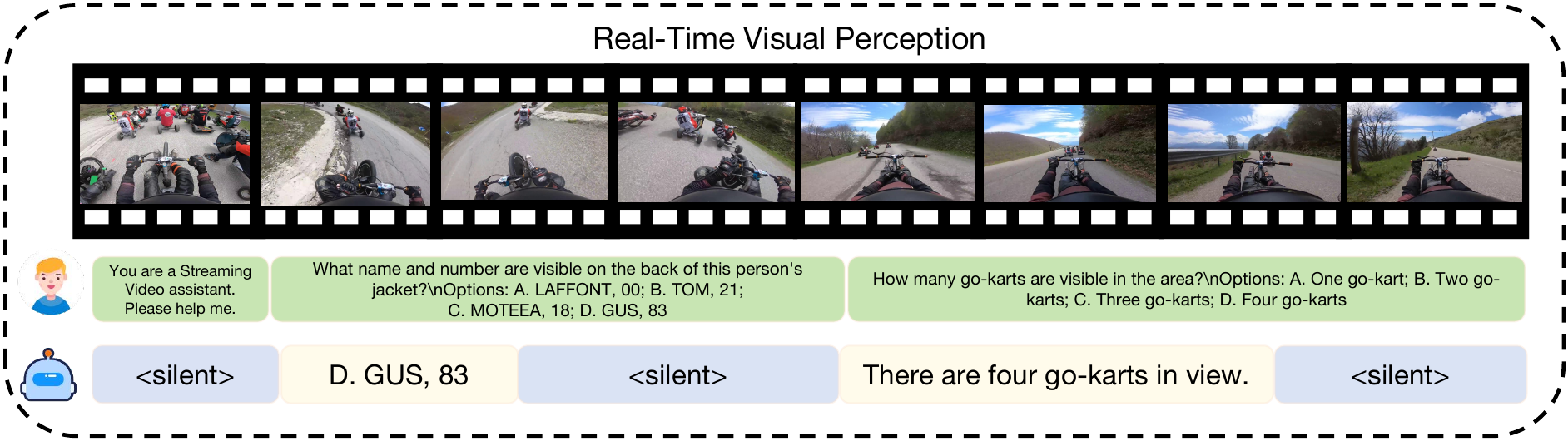}
    \caption{A case of Real-Time Visual Perception. The model provides immediate answers to visual queries about the current frame while remaining silent at other times.}
    \label{fig:case3}
\end{figure*}
We present qualitative examples illustrating EvoStreaming's behavior across three task categories.

\noindent \textbf{Forward Active Responding.} Figure~\ref{fig:case1} illustrates proactive anticipation in a cooking scenario. The model observes the soy milk preparation process and remains silent during routine steps. When the person transitions to a new step (pouring water, pressing buttons, filtering), the model generates concise descriptions. This behavior is consistent with learning to identify moments that contain actionable information while withholding responses during less informative intervals.

\noindent \textbf{Backward Tracing.} Figure~\ref{fig:case2} shows retrospective reasoning in an egocentric setting. When queried about past events (``Who did I talk to?'', ``Where was the phone?''), the model retrieves information from its accumulated history rather than current frames. The silent periods between queries suggest that the model can maintain relevant context without verbose output, then produce answers when explicitly asked.

\noindent \textbf{Real-Time Visual Perception.} Figure~\ref{fig:case3} illustrates immediate visual recognition. Upon receiving queries about visible content (jacket text, go-kart count), the model provides answers based on the current frame. The response timing suggests that the model can use current visual evidence without needing to aggregate evidence across multiple frames.

Collectively, these cases illustrate three interaction patterns: proactive anticipation for future events, retrospective retrieval from memory, and immediate response to present stimuli. The consistent use of \texttt{<silent>} tokens across these examples is consistent with learning a sparser response policy rather than simply generating continuous output.

\section{Broader Impacts}
\label{sec:broader_impacts}

Streaming video assistants could improve accessibility (e.g., live captioning and scene description), telepresence, and embodied systems where timely, sparse responses matter. RealStreamEval and verbosity-aware metrics encourage systems that speak only when useful, which may reduce distraction and energy use on wearable and edge devices compared to always-verbose models.

Potential negative impacts mirror those of any always-on video interface: continuous capture raises privacy and consent issues, and mistaken or mistimed outputs could mislead users in safety-critical settings. Stronger streaming models could also lower the barrier to large-scale passive monitoring if deployed without safeguards. We do not release a new foundation model; our method adapts existing open VideoLLMs, so misuse risk is largely inherited from upstream capabilities and deployment choices rather than introduced solely by this work.

\end{document}